\documentclass[sigconf]{acmart}
\makeatletter
\def\@ACM@checkaffil{
    \if@ACM@instpresent\else
    \ClassWarningNoLine{\@classname}{No institution present for an affiliation}%
    \fi
    \if@ACM@citypresent\else
    \ClassWarningNoLine{\@classname}{No city present for an affiliation}%
    \fi
    \if@ACM@countrypresent\else
        \ClassWarningNoLine{\@classname}{No country present for an affiliation}%
    \fi
}
\makeatother

\newcommand{\ie}{\textit{i}.\textit{e}., }
\newcommand{\eg}{\textit{e}.\textit{g}., }
\usepackage{algorithm}
\usepackage{listings}
\usepackage{color}
\usepackage{etoolbox}
\usepackage{multirow}
\usepackage{booktabs}
\usepackage{algpseudocode}
\usepackage{subfigure}

\makeatletter
\AfterEndEnvironment{algorithm}{\let\@algcomment\relax}
\AtEndEnvironment{algorithm}{\kern2pt\hrule\relax\vskip3pt\@algcomment}
\let\@algcomment\relax
\newcommand\algcomment[1]{\def\@algcomment{\footnotesize#1}}
\renewcommand\fs@ruled{\def\@fs@cfont{\bfseries}\let\@fs@capt\floatc@ruled
  \def\@fs@pre{\hrule height.8pt depth0pt \kern2pt}%
  \def\@fs@post{}%
  \def\@fs@mid{\kern2pt\hrule\kern2pt}%
  \let\@fs@iftopcapt\iftrue}
\makeatother

\AtBeginDocument{%
  }

\setcopyright{acmcopyright}
\copyrightyear{2023}
\acmYear{2023}
\acmDOI{XXXXXXX.XXXXXXX}

\acmConference[MM'23]{Make sure to enter the correct
  conference title from your rights confirmation emai}{Oct. 29--Nov. 3, 2023}{Ottawa, Canada}
\acmPrice{15.00}
\acmISBN{978-1-4503-XXXX-X/18/06}

\renewcommand\footnotetextcopyrightpermission[1]{}
\acmSubmissionID{1447}

\begin{document}

\title{Capturing Co-existing Distortions in User-Generated Content for No-reference Video Quality Assessment}


\author{Kun Yuan$^{\dag}$}
\email{yuankun03@kuaishou.com}
\affiliation{
  \institution{Kuaishou Technology}
}

\author{Zishang Kong$^{\dag}$}
\email{zishang.kong@pku.edu.cn}
\affiliation{
  \institution{Peking University}
}

\author{Chuanchuan Zheng}
\email{zhengchuanchuan@kuaishou.com}
\affiliation{
  \institution{Kuaishou Technology}
}

\author{Ming Sun}
\email{sunming03@kuaishou.com}
\affiliation{
  \institution{Kuaishou Technology}
}

\author{Xing Wen}
\email{wenxing@kuaishou.com}
\affiliation{
  \institution{Kuaishou Technology}
}


\begin{abstract}
Video Quality Assessment (VQA), which aims to predict the perceptual quality of a video, has attracted raising attention with the rapid development of streaming media technology, such as Facebook, TikTok, Kwai, and so on. Compared with other sequence-based visual tasks (\eg action recognition), VQA faces two under-estimated challenges unresolved in User Generated Content (UGC) videos. \textit{First}, it is not rare that several frames containing serious distortions (\eg blocking, blurriness), can determine the perceptual quality of the whole video, while other sequence-based tasks require more frames of equal importance for representations. \textit{Second}, the perceptual quality of a video exhibits a multi-distortion distribution, due to the differences in the duration and probability of occurrence for various distortions. In order to solve the above challenges, we propose \textit{Visual Quality Transformer (VQT)} to extract quality-related sparse features more efficiently. Methodologically, a Sparse Temporal Attention (STA) is proposed to sample keyframes by analyzing the temporal correlation between frames, which reduces the computational complexity from $O(T^2)$ to $O(T \log T)$. Structurally, a Multi-Pathway Temporal Network (MPTN) utilizes multiple STA modules with different degrees of sparsity in parallel, capturing co-existing distortions in a video. Experimentally, VQT demonstrates superior performance than many \textit{state-of-the-art} methods in three public no-reference VQA datasets. Furthermore, VQT shows better performance in four full-reference VQA datasets against widely-adopted industrial algorithms (\ie VMAF and AVQT).
\end{abstract}




\maketitle

\def\thefootnote{\dag}\footnotetext{Authors contributed equally to this research.}

\section{Introduction}


User Generated Content (UGC) has brought evolution to the daily-life consumer domain, which empowers amateurs to become active producers more than consumers. Lower video production cost leads to an explosion of UGC videos on video-sharing platforms, such as FaceBook, Kwai, and so on, which aims to deliver high-quality Quality of Experience (QoE) / Quality of Service (QoS) experience to users. 
Compared with Professionally Generated Content (PGC), UGC videos inevitably have worse conditions of shooting, poor capturing equipment, and unstable transmission links \cite{DBLP:conf/cvpr/WangKTYBAMY21}. 
For video streaming services, VQA has attracted more attention to filter out videos with low perceptual qualities \cite{DBLP:conf/cvpr/FangZZMW20,DBLP:journals/tcsv/GhadiyaramPB19}. Furthermore, VQA is also used to conduct content-aware video encoding \cite{DBLP:conf/cvpr/ChadhaA21} or enhancement \cite{DBLP:conf/iccp/TalebiM18,DBLP:conf/eccv/JohnsonAF16,DBLP:conf/cvpr/ZhangIESW18}, resulting in lower bandwidth cost and better viewing experience. Therefore, it is of great economic value to rate the perceptual quality of UGC videos through VQA.

\begin{figure}[t]
\begin{center}
\includegraphics[width=\linewidth]{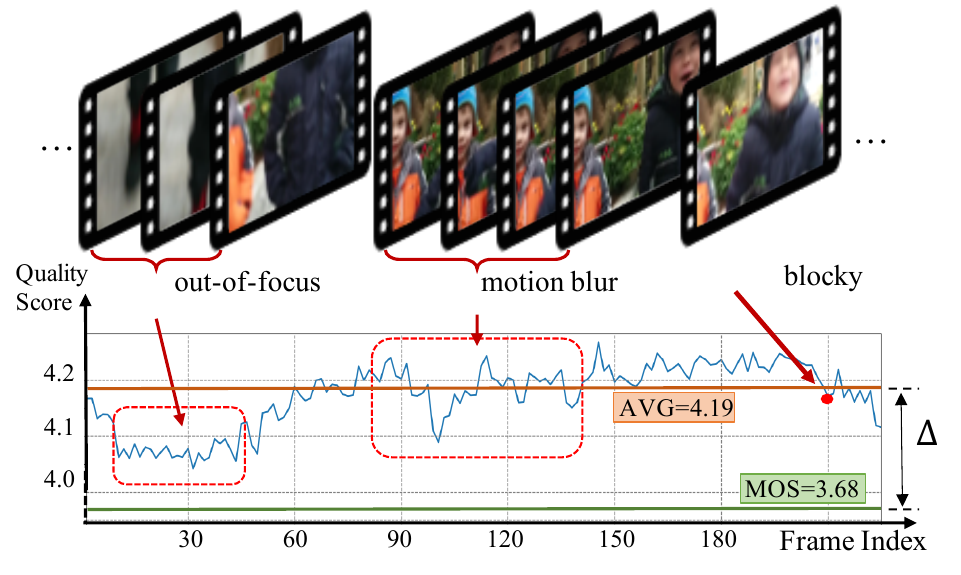}
\end{center}
  \caption{
  An example of UGC videos of No.13079468475 in the KoNViD-1k \cite{DBLP:conf/qomex/HosuHJLMSLS17}. This video contains multiple distortions along the temporal dimension, including motion-related types (\textit{i.e.} out-of-focus and blurriness) and compression-related types (\textit{i.e.} blocking artifacts). There exists a large difference in perceptual quality among frames. And those frames with low quality determine the overall quality instead of an arithmetic mean. Therefore, a more appropriate strategy is needed for spatiotemporal representation in VQA.
  }
\label{fig:video_example}
\end{figure}


A large number of studies on image/video quality assessment (QA) are studied in the previous literature. According to the \textit{availability of reference videos}, QA measures can be categorized into full-reference (FR) \cite{DBLP:journals/ijon/GaoWLTYZ17,DBLP:conf/cvpr/Kim017}, reduced-reference (RR) \cite{DBLP:journals/sigpro/MinGZHY18} and no-reference (NR) \cite{DBLP:journals/ijcv/LiJJ21,DBLP:journals/spl/MittalSB13,DBLP:conf/iccv/KeWWMY21,DBLP:journals/corr/abs-2206-09853}. Since distortion-free reference videos are often hard to obtain, NR-VQA is widely adopted in the UGC scenarios.
According to the \textit{feature generation types}, QA methods are divided into traditional hand-crafted \cite{DBLP:journals/tip/MittalMB12,DBLP:journals/spl/MittalSB13} and learning-based \cite{DBLP:conf/aaai/GuMDXP19,DBLP:conf/icip/BosseMWS16,DBLP:conf/cvpr/KangYLD14,DBLP:conf/icdsp/LiPFY16} ones. With the rapid development of deep learning, convolutional neural networks (CNN) \cite{DBLP:conf/mm/LiJJ19,DBLP:journals/ijcv/LiJJ21,DBLP:journals/tcsv/ZhangGHLH19} and Transformers \cite{DBLP:conf/iccv/KeWWMY21,DBLP:journals/corr/abs-2108-09635,DBLP:journals/corr/abs-2206-09853} are also used to boost the VQA domain.


As shown in Fig.~\ref{fig:video_example}, the common phenomenon is that multiple distortions co-occur within a UGC video, where different distortions begin to appear at different frames and own different time-span. Such a phenomenon casts two challenges for a better fit of human perceptual quality.
\textbf{First}, the perceptual quality of a video is determined by the keyframes that contain particular distortions. Excessively dense sampling brings an unbalanced distribution of frames and may disturb the learning process of distortion characteristics. While relatively current sparse sampling may ignore keyframes. How to select frames efficiently is an essential problem to be solved.
\textbf{Second}, due to the differences in temporal duration of different distortions, the perceptual quality of frames within a video exhibits a multi-distribution mode. Take some distorted characteristics for example, blocking artifact \cite{DBLP:conf/cvpr/KangYLD14}, dirty lens \cite{DBLP:journals/tog/GuRBN09}, and noise  \cite{DBLP:conf/bmsb/LiuZZSGY14,zhou2014no} are usually easy-detected given an individual frame. But out-of-focus and motion blurriness \cite{DBLP:journals/jvcir/LiuGZLZG17} can only be recognized using multiple frames. These factors put forward a higher request for VQA methods with the ability to perform frames analysis under different durations simultaneously. 

To overcome the aforementioned annoying challenges, we propose \textit{Visual Quality Transformer (VQT)} to extract quality-aware features focusing on multi-distortions more efficiently. Specifically, \textbf{to solve the first challenge}, a \textit{Sparse Temporal Attention (STA)} is proposed to sample keyframes via analyzing the temporal correlation between frames. It reformulates self-attention from the perspective of sparse sampling and adopts a proper sampling ratio according to the Johnson-Lindenstrauss (JL) lemma \cite{jllemma1984}. The keyframes can be selected by comparing the Kullback–Leibler (KL) difference between the Uniform distribution and its cosine similarity with other frames. As for model efficiency, compared with vanilla temporal attention, STA reduces the computational complexity from $O(T^2)$ to $O(T \log T)$. 
\textbf{To solve the second challenge}, owing to the efficiency of the STA module, \textit{Multi-Pathway Temporal Network (MPTN)} is adopted to capture co-existing distortions in a video simultaneously, which stacks multiple STA modules with different degrees of sparsity. Finally, the aggregated features are used for the representation of a video in VQA.

Our \textbf{contributions} are summarized as follows:
\begin{itemize}
    \item We propose an effective and efficient Visual Quality Transformer (VQT) for the NR-VQA tasks, where the proposed STA selects key frames containing particular distortions and the MPTN helps capture different distorted characteristics simultaneously in UGC scenarios. 
    \item VQT demonstrates superior performance than many \textit{state-of-the-art (SoTA)} methods in three NR-VQA datasets, raising the performance by 2.14$\%$ of PLCC in KoNViD-1k and 2.17$\%$ of PLCC in YouTube-UGC over the best results. Furthermore, VQT obtains better results in four FR-VQA datasets (cross-dataset evaluation in three of them) against widely-adopted industrial algorithms (\eg VMAF and AVQT).
    \item VQT can act as a plug-in module used for general computer-vision tasks and shows good generalization ability in the video classification task. Compared with the original dense attention mechanism (\eg TimeSformer), the computational cost decreases from 197 TFLOPs to 154 TFLOPs (-22\%).
\end{itemize}

\section{Related Work}
\subsection{Perceptual Quality Assessment}

According to the accessibility of the reference images or videos, QA is divided into FR-QA, RR-QA, and NR-QA tasks. FR-QA and RR-QA tasks require a full and partial reference respectively. While the NR-QA method only takes distorted images or videos as input, which is often more challenging, but also more practical in most scenarios. In this paper, we focus on the NR-VQA domain.

In the early period of NR-VQA, most works \cite{DBLP:journals/tcsv/AmerD05,DBLP:conf/icip/MarzilianoDWE02,DBLP:conf/icip/WangBE00} focused on identifying specific types of distortions(\eg blur, blocky). Then, more methods \cite{DBLP:conf/cvpr/YeKKD12,DBLP:journals/tip/MittalMB12} have been proposed to focus on multiple distortions jointly to carry out comprehensive QA.
With the rapid progress of deep learning, learning-based methods \cite{DBLP:journals/corr/abs-2101-10955,DBLP:conf/eccv/KimKAKL18,DBLP:conf/cvpr/YingMGB21,DBLP:journals/tcsv/ZhangGHLH19,Zhao_2023_CVPR,Zhao_2023_CVPR_Zoom} have suppressed the performance of traditional hand-crafted ones, due to their versatility and generalization.
RAPIQUE \cite{DBLP:journals/corr/abs-2101-10955} combined quality-aware features of scene statistics and semantics-aware deep convolutional features. 
A combination of 3D-CNN and LSTM was adopted \cite{DBLP:conf/icip/YouK19} to extract local spatiotemporal features from small clips in the video.
Patch-VQA \cite{DBLP:conf/cvpr/YingMGB21} devised a local-to-global patch-based architecture, and extracted both 2D and 3D video features using a temporal DNN to predict the quality.
STDAM \cite{DBLP:conf/mm/XuLZZW021} introduced using the graph convolution and attention module to extract and enhance the quality-related features. 2BiVQA \cite{telili20222bivqa} proposed using two Bi-directional Long Short Term Memory (Bi-LSTM) to conduct the quality assessment. One is for capturing short-range dependencies between image patches, and the other is for capturing long-range dependencies between frames.
However, how to efficiently select keyframes to model quality-related features is still an open problem in the VQA domain. Recently, DisCoVQA \cite{DBLP:journals/corr/abs-2206-09853} designs a Transformer-based Spatial-Temporal Distortion Extraction module to tackle temporal quality attention.

\subsection{Video Transformer Architecture}

Visual Transformers \cite{DBLP:conf/iclr/DosovitskiyB0WZ21} are the most popular alternatives applied in various vision tasks, due to their good ability in modeling long-term dependency of sequential data. Following ViT, many Transformer-based models \cite{DBLP:conf/iccv/Arnab0H0LS21,DBLP:conf/icml/BertasiusWT21,Fan_2021_ICCV,DBLP:journals/corr/abs-2106-13230,Yuan_2021_ICCV} were developed for video classification tasks. TimeSformer \cite{DBLP:conf/icml/BertasiusWT21} explored the factorization of spatial-temporal dimension for efficient computation. Video Swin Transformer \cite{DBLP:journals/corr/abs-2106-13230} globally connected patches across the spatial and temporal dimensions, and advocated an inductive bias of locality for a better speed-accuracy trade-off. There also exists some work exploring the applicability of Transformers in the field of VQA. B-VQA \cite{DBLP:journals/corr/abs-2108-08505} combined GRU unit with Transformer encoder to model the temporal information, which further boosts the performance. StarVQA \cite{DBLP:journals/corr/abs-2108-09635} transferred the divided space-time attention in TimeSformer directly into VQA, showing well generalization ability in the regression task. But these architectures are not designed specifically for VQA, especially for efficiently modeling co-existing distortions. In this paper, we deeply analyze the problems faced by VQA and design the VQT in a targeted manner.

\section{Method}

\begin{figure*}[t]
\begin{center}
\includegraphics[width=0.9\textwidth]{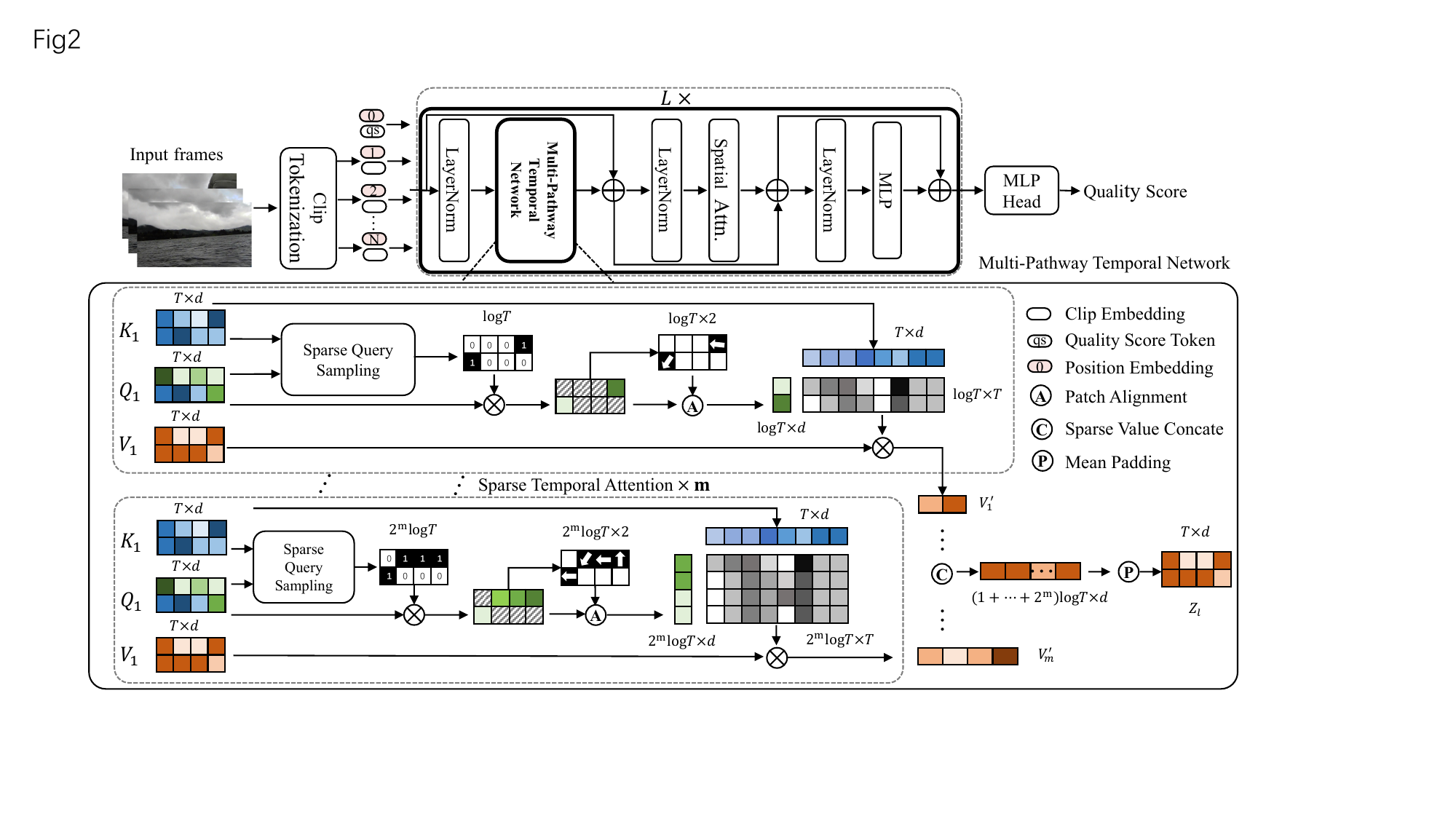}
\end{center}
\caption{
Illustration of the proposed Visual Quality Transformer (VQT). It receives a clip as input and reshapes each frame into patches for tokenization. 
Then the sequence of tokens is fed into a stacked encoder, performing the spatiotemporal attention. 
In the temporal dimension, VQT utilizes a Multi-Pathway Temporal Network to capture different distorted characteristics simultaneously with stacking Sparse Temporal Attention blocks.
In each block, STA conducts sparse query sampling to locate keyframes for distortion representation. 
To further enhance STA with spatial relationships, temporal offsets across frames are assigned to patches for alignment. 
Different blocks contribute to the final representation within an encoder block. To predict the video quality, the corresponding quality score token is used for the final representation.
}
\label{fig:VQT}
\end{figure*}

\subsection{Revisiting Video Transformer}

Different from image Transformers, video Transformers receive a sequence of frames as input. There exist some differences in basic architecture modules, including clip tokenization and spatiotemporal attention. We first briefly revisit them for a better understanding.

\paragraph{Clip Tokenization} 
The video Transformer takes a clip as input denoted as $\mathbf{X} \in \mathbb{R}^{T \times H\times W \times 3}$, which composes of $T$ RGB frames with the size of $H \times W$ under equal temporal interval sampled from original videos.
Following ViT, video Transformers reshape each frame into $N$ non-overlapping patches, where each size is $P\times P$ and $N = \frac{H \times W}{P^2}$. Then the sequence of frames can be flattened into $\mathbf{X} \in \mathbb{R}^{T \times N \times (3P^2)}$.
Besides, an extra learnable positional embedding $\mathbf{E}^{p o s}$ is added to encode the spatiotemporal position of each patch. Then the input embedding $\mathbf{Z} \in \mathbb{R}^{T \times N \times d}$ is calculated as:
\begin{equation}\label{eq1}
    \mathbf{Z}=\mathbf{W} \mathbf{X}^{\top}+\mathbf{E}^{p o s},
\end{equation}
where $\mathbf{W} \in \mathbb{R}^{d \times 3 P^{2}}$ is the mapping weight and $d$ is the embedding dimension of each frame token.

\paragraph{Divided Space-Time Attention} 
To process spatiotemporal information, TimeSformer utilizes the ``Divided Space-Time Attention" module, where the temporal attention and the spatial attention are performed sequentially. In each encoder block $\ell$, the temporal attention first computes the relationship among all patches in the same spatial location from different frames, expressed as:
\begin{equation}
\label{eq2}
    \mathbf{Z}^{\ell}_{\mathrm{time}} = \mbox{Softmax}\left(\frac{\mathbf{Q}_{t^{\prime}}^{\ell - 1}}{\sqrt{d}}{ \mathbf{K}^{\ell - 1}_{t^{\prime}}}^{\top}\right) \mathbf{V}^{\ell - 1},
\end{equation}
where $\mathbf{Q}$, $\mathbf{K}$, $\mathbf{V}$ are the query, key, and value. And $t^{\prime}$ denotes that the inner products are computed on the temporal dimension. Then the generated $\mathbf{Z}^{\ell}_{\mathrm{time}}$ is fed back for the spatial attention, computing as:
\begin{equation}\label{eq3}
\mathbf{Z}^{\ell}{\mathrm{space}} = \mathrm{Softmax}\left(\frac{\mathbf{Q}_{p^{\prime}}^{\ell - 1}}{\sqrt{d}} \mathbf{K}^{\ell - 1}_{p^{\prime}}{}^{\top}\right) \mathbf{V}^{\ell - 1},
\end{equation}
where $p^{\prime}$ denotes inner products on the spatial dimension.

\subsection{Visual Quality Transformer}

The illustration of VQT is given in Fig.~\ref{fig:VQT}. The core components are STA and MPTN, which we will describe in more detail.

\begin{figure}[t]
\centering
\subfigure{
    \begin{minipage}[b]{0.9\linewidth}
    \includegraphics[width=\linewidth]{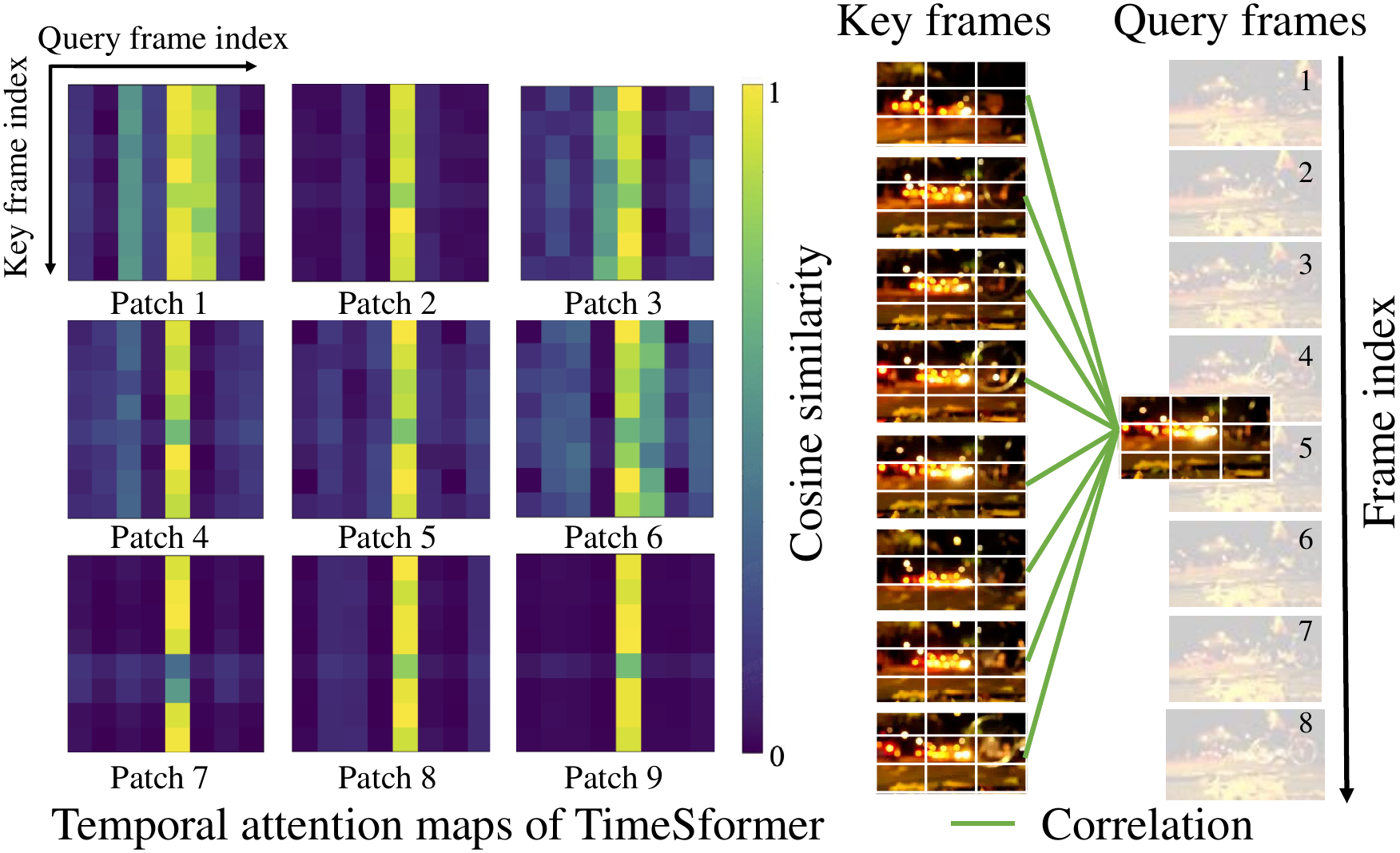}
    \end{minipage}
}
\subfigure{
    \begin{minipage}[b]{0.45\linewidth}
    \includegraphics[width=\linewidth]{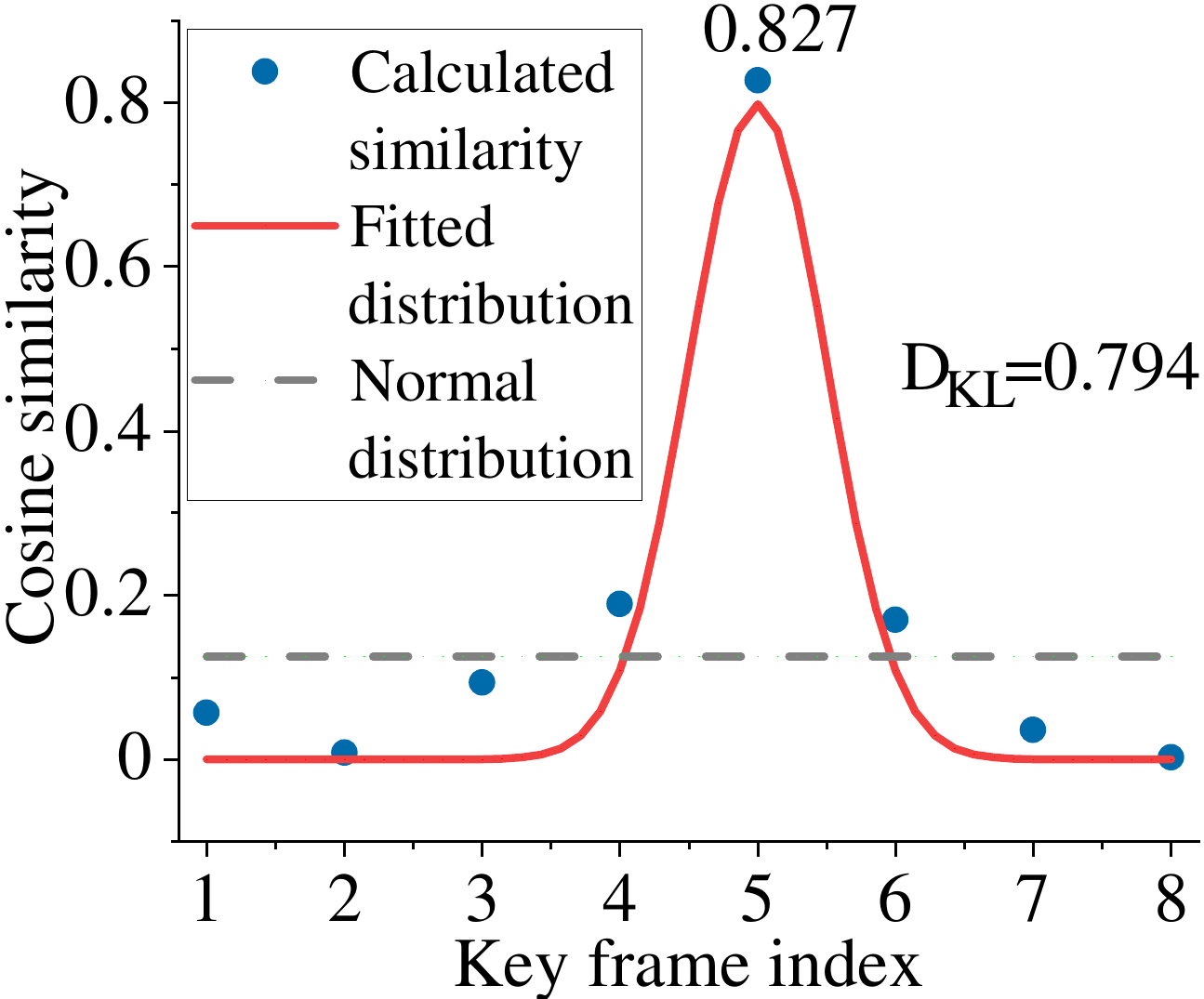}
    \end{minipage}
}
\subfigure{
    \begin{minipage}[b]{0.45\linewidth}
    \includegraphics[width=\linewidth]{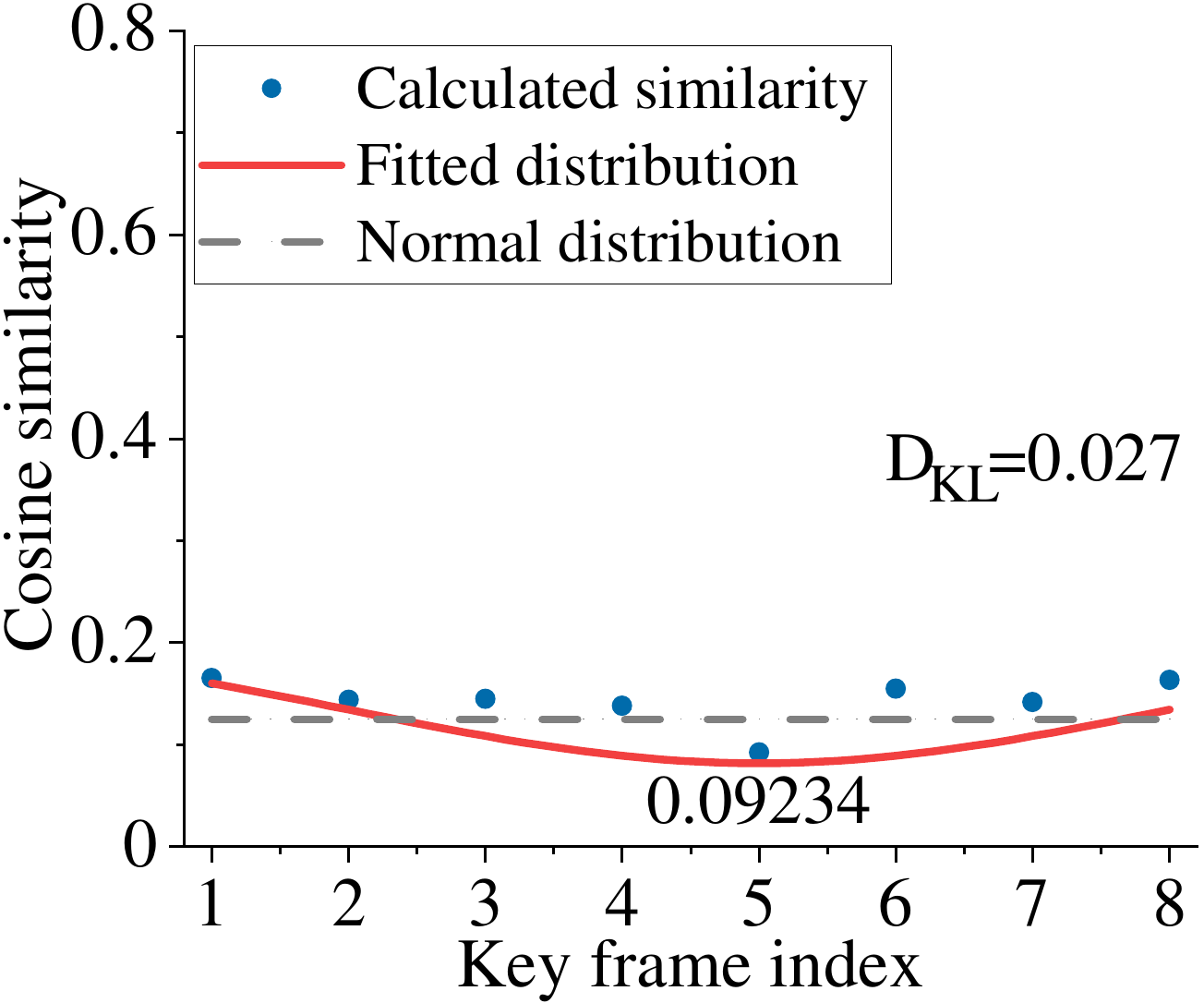}
    \end{minipage}
}
\caption{The upper images show that the quality of the video is largely affected by frames containing distortions. The below curves give the distributions of $D_{KL}$ for the 5-\textit{th} (left) and 3-\textit{th} (right) frame.}
\label{fig:visulization}
\end{figure}

\paragraph{Sparse Temporal Attention} 
Video clips own widespread information redundancy in the temporal dimension \cite{DBLP:conf/cvpr/WangLWY17,DBLP:journals/corr/abs-2006-15489}, both in the frame level and feature level. Frames containing distorted characteristics largely reflect the perceptual quality of the whole video. The upper part of Fig. \ref{fig:visulization} demonstrates this phenomenon in the VQA domain, where the computed temporal attention map shows that all frames have a strong correlation with the 5-$th$ frame (\ie, an over-exposed distorted image). And the factor of overexposure in this frame deteriorates the quality of this video. Based on the observations in Fig.~\ref{fig:video_example} and Fig.~\ref{fig:visulization}, we propose a Sparse Temporal Attention, aiming to sample key-frames containing distortions in a video.

Specifically, the attention mechanism is first reformatted from the perspective of sparse sampling. Given the query $\mathbf{Q} \in \mathbb{R}^{T \times d}$, key $\mathbf{K} \in \mathbb{R}^{T \times d}$, and value $\mathbf{V} \in \mathbb{R}^{T \times d}$, the self-attention is computed as $\mathbf{V}^\prime = \mbox{softmax}(\frac{\mathbf{QK^\top}}{\sqrt{d}})\mathbf{V}$, where $\mathbf{V}^\prime \in \mathbb{R}^{T \times d}$.
According to the JL lemma, \textit{there exists a linear transformation that projects $\mathbf{V}$ into $\mathbf{V}^{\prime} \in \mathbb{R}^{\log T \times d}$ with low-distortion embedding}, where $\log T$ is the minimal number of frames sampled from $T$ frames.
Specifically, we let $q_i$ and $k_j$ represent the feature of $i$-\textit{th} row and $j$-\textit{th} row in $\mathbf{Q}$ and $\mathbf{K}$, respectively. Then we calculate the cosine similarity between $q_i$ and $k_{j, j=1,..,T}$. And the computed distribution can be noted as $p(\mathbf{K} |q_i)$, representing the correlation of the $i$-\textit{th} frame with all frames. Then we compute the KL divergence between $p(\mathbf{K} |q_i)$ and the uniform distribution $U$:
\begin{align}
    D_{\mathrm{KL}}( p(\mathbf{K}|q_i) \| U) = \sum_{j=1}^{T} \frac{1}{T} \log \frac{1}{T}-\frac{1}{T} \log \left(\frac{q_{i} k_{j}^{T}}{\sum_{j=1}^{T} q_{i} k_{j}^{T}}\right) \\
    = -\log \frac{1}{T}-\sum_{i=1}^{T} \frac{1}{T}\left(\ln \exp \frac{q_{i} k_{j}^{T}}{\sqrt{d}}-\ln \sum_{j=1}^T e^{\frac{q_{i} k_{j}^{T}}{\sqrt{d}}}\right).
\end{align}
For simplicity, we ignore the constant term of $-\log \frac{1}{T}$. And the equation can be rewritten as:
\begin{equation}
        \mbox{D}_{KL}(p(\mathbf{K}|q_i)|| U) = \ln \sum_{j=1}^{T} e^{\frac{{q}_{i} {k}_{j}^{\top}}{\sqrt{d}}}-\frac{1}{T}\sum_{j=1}^{T}\left(\frac{{q}_{i} {k}_{j}^{\top}}{\sqrt{d}}\right).
\end{equation}
Larger values of $\mbox{D}_{KL}(p(\mathbf{K}|q_i)|| U)$ indicate a \textbf{stronger correlation with other frames}, which can be used for the selection of keyframes. Examples are given in Fig. \ref{fig:visulization}, where frames containing distortions own larger (\eg, the 5-$th$ frame) values and vice versa.

To reduce the complexity of computing the divergence of all frames, $\log T$ frames are sampled randomly. The distribution of divergence of all frames can be estimated by calculating the mean $\mu$ and variance $\sigma$ of $\log T$ frames. Then $\log T$ query frames can be selected, whose divergence meets $\mbox{D}_{KL}(p(\mathbf{K}|q_i)|| U) > \mu + \sigma$. The dimension of sampled query $\mathbf{\hat{Q}}$ is reduced to $\log T \times d$. And the pseudocode is shown in Alg.~1.

To enhance the spatial relationship of queries between different frames, STA further performs a spatial shift for alignment. Specifically, we reshape the spatial dimension of selected query features $\mathbf{\hat{Q}}\in \mathbb{R}^{\log T\times N \times d }$ into $ \mathbb{R}^{\log T\times \sqrt{N} \times \sqrt{N} \times d}$ (the spatial dimension $N$ is not shown in Fig. \ref{fig:VQT} for simplify). Then a linear projection layer is attached on the reshaped features to predict 2D offsets $\mathbf{P}$ for each token, where $\mathbf{P}\in \mathbb{R}^{\log T\times \sqrt{N} \times \sqrt{N} \times 2}$. Then the shifted query features $\mathbf{\hat{Q}}^{\prime}$ can be obtained by bilinear interpolation. Then the weighted value features can be computed by:
\begin{equation}
    \mathbf{V}^\prime = \mbox{softmax}(\frac{\mathbf{\mathbf{\hat{Q}}^{\prime} K^\top}}{\sqrt{d}})\mathbf{V}.
\end{equation}

\paragraph{Multi-Pathway Temporal Network} 
To capture different distorted characteristics simultaneously, multiple STA modules with different degrees of sparsity are stacked in parallel.
Given a video clip with $T$ frames, the number $m$ of pathway is determined by $m = \lfloor \log ( \frac{T}{\lceil \log T \rceil}+1) \rfloor - 1$, where $\lceil \cdot \rceil$ and $\lfloor \cdot \rfloor $ represents the ceiling and flooring operation respectively.
The minimal number of frames for different STA modules is $\log T$, and the maximal is $2^m \log T$. Take an input clip with 96 frames, for example, its $m$ is 3, containing 3 pathways. And each pathway selects 7, 14, and 28 keyframes respectively. Each block performs temporal attention over selected frames, resulting in a weighted value of $\mathbf{V}^{\prime}_m \in \mathbb{R}^{2^m \log T \times d}$. Then the values generated by different blocks are concatenated in the temporal dimension, resulting in $\hat{\mathbf{V}} \in \mathbb{R}^{(1 + \cdots + 2^m) \log T \times d}$. To align with succeeding encoder blocks, we use the mean padding operation to fill $\hat{\mathbf{V}}$ in the temporal dimension, generating $\mathbf{Z}_l \in \mathbb{R}^{T\times d}$:


\begin{algorithm}[t]
\label{alg:code}
	\caption{Pseudo-code of selection of keyframes} 
	\begin{algorithmic}[1]
        \State $Q_{sample}=random.sample(Q, gamma)$ \Comment{gamma:logT/T}
        \State $mu=KL$-$Divergence(Q_{sample}, K).mean(dim=-1)$
        \State $sigma=KL$-$Divergence(Q_{sample}, K).std(dim=-1)$
        \State init $i=0$; $attn=zeros(logT, T)$
    \For {$q ~in ~Q$}
        \If {$i<logT$ and $KL$-$Divergence(q,K)>mu+sigma$}
            \State $attn[i,:]=q@K.transpose()$
            \State $i$ +=$1$
            \EndIf
    \EndFor
        \State $V=bmm(attn, V)$ \Comment{bmm:batch matrix multiplication}
	\end{algorithmic} 
\end{algorithm}

\begin{equation}\label{stacking}
    \begin{aligned}
    \hat{\mathbf{V}} &=\mbox{Concate}(\mathbf{V}^{\prime}_1, \cdots, \mathbf{V}^{\prime}_m) \\
    \mathbf{Z}_{l} & =\mbox{Mean-Padding}(\hat{\mathbf{V}}).
    \end{aligned}
\end{equation}


\subsection{Optimization Objective}


A smooth $\mathcal{L}_1$ loss is adopted to train VQT models.
Let $\mathcal{F}(\cdot)$ represent the mapping function of the VQT model. Given mini-batch videos during training, the objective function can be denoted as: 
\begin{equation}
\begin{aligned}
    \mbox{min} \enspace \frac{1}{|\mathcal{B}|} \sum_{i=1}^{\mathcal{B}} \mathcal{L}_{1-smooth} & (\mathcal{F}(\mathbf{X}_i),y_i),
\end{aligned}
\end{equation}
where $\mathbf{X}_i$ and $y_i$ is the input video and corresponding labeled Mean Opinion Score (MOS). $\mathcal{B}$ indicates the size of the mini-batch.

\subsection{Computational Efficiency}

\paragraph{Computational Complexity of STA} For an input clip with $T$ frames, the computational cost of the original temporal attention module is $O(T^2)$. The STA module applies sparse computation among the temporal dimension with a cost of $O(2^{m}\log T\cdot T)$. And $2^{m}\log T$ is smaller than $T$ as mentioned above. 
\paragraph{Computational Complexity of MPTN} MPTN is composed of multiple STA blocks, whose total computational cost is computed by combining each one of $O(\sum_{a=0}^{m} 2^a \log T \cdot T \cdot d)$.  Since $\sum_{a=0}^{m} 2^a \log T$ is smaller than $T$, the computational cost of MPTN is still less than that of the original temporal attention module.
\paragraph{Measured Inference Speed} Further evaluations of efficiency are given in the following experiments. In Tab.~\ref{tab:speed} and ~\ref{tab:k400}, compared with the original version of dense attention (\ie, TimeSformer used in StarVQA), VQT has a less computational cost (-22\%), faster inference speed (+13\%) and higher performance in PLCC (+3.49\%).




\section{Experiments}

\subsection{Datasets and Evaluation}

\paragraph{NR datasets.} 
We leverage three NR-VQA datasets to evaluate VQT models: LIVE Video Quality Challenge Database (LIVE-VQC) \cite{DBLP:journals/tip/SinnoB19}, Konstanz Natural Video Database (KoNViD-1k) \cite{DBLP:conf/qomex/HosuHJLMSLS17}, and Blind Video Quality Assessment for User Generated Content (Youtube-UGC) \cite{DBLP:conf/mmsp/WangIA19}. Subjective quality scores are provided in the form of MOS.
LIVE-VQC contains 585 videos labeled by MOS with a resolution from 240P to 1080P. 
KoNViD-1k comprises a total of 1,200 videos with a resolution of $960\times540$ that are fairly sampled from a large public video dataset, YFCC100M\cite{DBLP:journals/cacm/ThomeeSFENPBL16}. The duration of videos is 8s with 24/25/30FPS, whose MOS ranges from 1.22 to 4.64.
Youtube-UGC is composed of 1,500 videos that are sampled from millions of YouTube videos belonging to 15 categories annotated by a knowledge graph. The resolutions of videos are from 360P to 4K. For the Youtube-UGC, we follow the default training and testing splits \cite{DBLP:conf/mmsp/WangIA19}. For LIVE-VQC and KoNViD-1k, following \cite{DBLP:journals/corr/abs-2101-10955}, $80\%$ of the dataset is used for training, and the remaining $20\%$ is used for testing. 


\paragraph{FR Datasets.} To verify the generalization ability to video codec field, four more FR-VQA datasets are tested, including ICME-FR \footnote{http://2021.ieeeicme.org/2021.ieeeicme.org}, VQEG HD3 \footnote{https://www.cdvl.org}, NFLX Video dataset \footnote{https://github.com/Netflix/vmaf} and the Waterloo IVC 4k Video Quality database \cite{waterloo}. The VQEG HD3 database is composed of 9 source clips with a resolution of 1080P. And the source clips are encoded into 63 distorted videos for evaluation. The NFLX Video consists of 34 source clips, whose duration is 6s. They are sampled from popular TV shows and movies on Netflix. Source clips are encoded at resolutions ranging from $384\times288$ to $1920\times1080$, resulting in about 300 distorted videos. The Waterloo IVC 4k Video Quality database is created from 20 pristine 4K videos. Each video is encoded by five encoders: HEVC, H264, VP9, AV1, and AVS2, and divided into three solutions ($960\times540$, $1920\times1080$ and $3840\times2160$ with four distortion levels, resulting in 1,200 encoded videos.

\paragraph{Evaluation Criteria.} Pearson’s Linear Correlation Coefficient (PLCC),  Spearman’s Rank-Order Correlation Coefficient (SROCC), Kendall’s Rank-Order Correlation Coefficient (KROCC), and Root Mean Square Error (RMSE) are used for evaluation. PLCC and RMSE measure the prediction accuracy, SROCC, and KROCC indicate the prediction monotonicity. Better VQA methods should have larger PLCC/SROCC/KROCC and smaller RMSE values.

\subsection{Implementation Details}
Our implementation is based on PyTorch \cite{DBLP:conf/nips/PaszkeGMLBCKLGA19} and MMAction2 \cite{2020mmaction2}. All models are trained using 4 NVIDIA Tesla V100. The number of encoders follows the original setting of TimeSformer. We set a patch size of 16 in clip tokenization. The embedding dimension $d$ is 768. We use models that have been pre-trained on ImageNet \cite{imagenet} and Kinetics-400 \cite{kinetics} for training. During the optimization procedure, we use the AdamW optimizer with a learning rate of 1e-5 decayed by a factor of 0.1 every 30 epochs, minimizing the $\mathcal{L}1$ loss. All models are trained for 90 epochs. By default,  the checkpoint generated by the last iteration is used for evaluation. The batch size of video clips is set to 4 with a clip length of 96. Other training settings are the same with \cite{DBLP:conf/icml/BertasiusWT21}. The median result of 10 repeat runs is used for Tab.~\ref{tab:konvid1k&LIVEVQC&YoutuUGC} with different random splits.

\subsection{Comparison with SoTA Methods}

\begin{table*}[t]
    \centering
    \caption{Quantitative results of different methods on three public NR-VQA datasets. Larger PLCC and SROCC indicate better performance. Besides, weighted average scores are reported based on the number
    of videos of three datasets. The best and second best performances are \textbf{highlighted} and \underline{underlined}. The mark ``-" denotes that results are not reported originally. The ``*" mark indicates using extra training data for QA tasks. The VQT models outperform almost all SoTA methods by large margins.}
    \begin{tabular}{c|cc|cc|cc|cc}
    \toprule
        \multirow{2}{*}{Method} & \multicolumn{2}{c|}{KoNViD-1k} & \multicolumn{2}{c|}{LIVE-VQC}  &\multicolumn{2}{c|}{Youtube-UGC} & \multicolumn{2}{c}{Weighted Average} \\
         & PLCC $\uparrow$ & SROCC $\uparrow$ & PLCC $\uparrow$ & SROCC $\uparrow$ & PLCC $\uparrow$ & SROCC $\uparrow$ & PLCC $\uparrow$ & SROCC $\uparrow$ \\
    \midrule
        VIIDEO~\cite{DBLP:journals/tip/MittalSB16}   & 0.303 & 0.298 & 0.2164 & 0.0332 & 0.1534  & 0.0580 & 0.2230 & 0.1459 \\
        NIQE~\cite{DBLP:journals/spl/MittalSB13}  & 0.5530 & 0.5417 & 0.6286 & 0.5957 & 0.2776 & 0.2379 & 0.4500 & 0.4225 \\
        CORNIA~\cite{DBLP:conf/cvpr/YeKKD12}  & 0.608 & 0.610 & - & - & - & - & - & -\\
        \hline
        BRISQUE~\cite{DBLP:journals/tip/MittalMB12}   & 0.626 & 0.654 & 0.638 & 0.592 & 0.395 & 0.382 & 0.5299 & 0.5265 \\
        VBLIINDS~\cite{DBLP:journals/tip/SaadBC14}  & 0.6576 & 0.6947 & 0.7120 & 0.7015 & 0.5551 & 0.5590 & 0.6431 & 0.6427 \\
        GRU-VQA~\cite{DBLP:conf/mm/LiJJ19}  & 0.744 & 0.755 & - & - & - & - & - & - \\
        TLVQM~\cite{DBLP:conf/mmsp/EbenezerSWWB20}   & 0.7688 & 0.7729 & 0.8025 & 0.7988 & 0.6590 & 0.6693 & 0.7284 & 0.7337 \\
        MDTVSFA~\cite{DBLP:journals/ijcv/LiJJ21}  & 0.7856 & 0.7812 & 0.7728 & 0.7382 & - & - & - & -\\
        UGC-VQA~\cite{DBLP:journals/tip/TuWBAB21}  & 0.7803 & 0.7832 & 0.7514 & 0.7522 & 0.7733 & 0.7787 & 0.7719 & 0.7754 \\
        PVQ~\cite{DBLP:conf/cvpr/YingMGB21}  & 0.786 & 0.791 & 0.837 & 0.827 & - & - & - & -\\
        RAPIQUE~\cite{DBLP:journals/corr/abs-2101-10955}  & 0.8175 & 0.8031 & 0.7863 & 0.7548 & 0.7684 & 0.7591 & 0.7907 & 0.7753 \\ 
        StarVQA~\cite{DBLP:journals/corr/abs-2108-09635}  & 0.796 & 0.812  & 0.808& 0.732 & - & - & - & - \\
        BVQA$^{*}$~\cite{DBLP:journals/corr/abs-2108-08505}& 0.8335 & 0.8362 & \textbf{0.8415} & \textbf{0.8412} & 0.8194  & 0.8312 & 0.8290 & \underline{0.8350} \\
        STDAM~\cite{DBLP:conf/mm/XuLZZW021}  & \underline{0.8415} & \underline{0.8448}  & 0.8204 & 0.7931 & \underline{0.8297} & \underline{0.8341} & \underline{0.8320} & 0.8305 \\
        2BiVQA~\cite{telili20222bivqa} & 0.835 & 0.815 & 0.832 & 0.761 & 0.790 & 0.771 & 0.794 & 0.800 \\
        DisCoVQA~\cite{DBLP:journals/corr/abs-2206-09853} & 0.847 & 0.847 & 0.826 & 0.820 & - & - & - & - \\
    \midrule
        {Our TimeSformer} & 0.8293 & 0.8342 & 0.8017& 0.7845 & 0.8279  & 0.8133 & 0.8235  & 0.8159 \\
        {VQT} & \textbf{0.8684} & \textbf{0.8582} & \underline{0.8357} & \underline{0.8238} & \textbf{0.8514} & \textbf{0.8357} & \textbf{0.8529} & \textbf{0.8421} \\
    \bottomrule
    \end{tabular}
    \label{tab:konvid1k&LIVEVQC&YoutuUGC}
\end{table*}

Extensive experiments are conducted to compare with SoTA QA methods. As given in Tab.~\ref{tab:konvid1k&LIVEVQC&YoutuUGC}, we report the PLCC and SROCC performance in KoNViD-1k, LIVE-VQC and Youtube-UGC datasets. Besides, weighted average scores are reported based on the number of videos of three datasets. Some observations and conclusions can be found here. 
\textit{First}, compared with IQA methods (i.e. NIQE \cite{DBLP:journals/spl/MittalSB13},  BRISQUE \cite{DBLP:journals/tip/MittalMB12}, CORNIA \cite{DBLP:conf/cvpr/YeKKD12}), VQT obtains a large performance lead (+31.54\%, +24.24\%, +26.04\% of PLCC in KoNViD-1k), showing the effectiveness of fusion strategy of frames over an arithmetic mean.
\textit{Second}, compared with hand-crafted features (i.e. \cite{DBLP:journals/tip/MittalSB16}), VQT demonstrates the utility of learning-based methods over prior knowledge. 
\textit{Third}, compared with CNN models (i.e. \cite{DBLP:journals/corr/abs-2108-08505,DBLP:journals/corr/abs-2101-10955,DBLP:conf/mm/XuLZZW021,DBLP:conf/cvpr/YingMGB21}), VQT shows the advantage of Transformer models in building long-range dependencies.
\textit{Fourth}, compared with current Transformer models (i.e. StarVQA \cite{DBLP:journals/corr/abs-2108-09635}, TimeSformer), VQT also evaluate the gains from architecture modification by large margins (+7.24\% of PLCC in KoNViD-1k, +2.77\% of PLCC in LIVE-VQC). Compared with current SOTA method, VQT also outperforms STDAM \cite{DBLP:conf/mm/XuLZZW021} in three datasets(+2.59\% of PLCC in KoNViD-1k, +1.53\% of PLCC in LIVE-VQC, and +2.17\% of PLCC in Youtube-UGC). Since the size of LIVE-VQC is the smallest (only containing 585 videos), BVQA \cite{DBLP:journals/corr/abs-2108-08505} achieves the highest performance by introducing extra QA training data. However, VQT still surpasses it in the weighted scores (+2.39\% of PLCC), showing strong generalization ability. Compared with recent DisCoVQA, VQT obtain a higher result of PLCC by 2.14\%.



\subsection{Comparing with VMAF/AVQT}

We further compare the effectiveness of our method with the two industrial standards, i.e. Netflix's VMAF \footnote{https://github.com/Netflix/vmaf} and Apple's AVQT \footnote{https://developer.apple.com/videos/play/wwdc2021/10145} on the four widely adopted open datasets. The VMAF and AVQT have been used as standard deals to their simplicity in computations and consistent performance across different types of videos. As shown in Tab.~\ref{tab:vmaf}, their performance is consistently high across the four datasets except for the \textit{Waterloo IVC 4k} since both VMAF and AVQT were developed before 4K videos were becoming popular. The performance decrease on \textit{Waterloo IVC 4k} indicates that the generability of these two algorithms is not always satisfying when a new video format is induced. 

We train our VQT model with data solely from the ICME-FR train split. However, we not only test it on the ICME-FR test split but also directly evaluate its performance on the aforementioned three datasets without any fine-tuning (\emph{i.e.}, \textbf{cross-dataset evaluation}). From Tab.~\ref{tab:vmaf}, we observe that our proposed VQT method further improves the performance. This enhancement is not only attributed to the fact that our model is learned from a large-scale dataset but also thanks to our carefully designed architecture that can effectively extract and integrate spatiotemporal features to better model users' Mean Opinion Scores (MOS). It is worth noting that our method is efficient and can be easily integrated and substituted by existing methods such as VMAF/AVQT for evaluating the performance of different video encoding strategies.

\begin{table}[t]
    \centering
    \caption{Comparisons with industrial standards. The VQT model is trained on the ICME-FR datasets with the supervision of DMOS. \textbf{And direct inference results are reported on the other three datasets without fine-tuning}. VQT shows strong generalization ability and practical prospects.}
    \begin{tabular}{c|c|cc}
    \toprule
        Datasets & Method & PLCC $\uparrow$ & SROCC $\uparrow$ \\
    \midrule
        \multirow{3}{*}{ICME-FR}& VMAF & 0.9423 & 0.9137 \\
        & AVQT & 0.9730 & 0.9334 \\
        & VQT  & \textbf{0.9867} & \textbf{0.9364} \\
    \midrule
        \multirow{3}{*}{NFLX Video Dataset}& VMAF  & 0.9351 & 0.9173 \\
        & AVQT & 0.9571  & 0.9420 \\
        & VQT  & \textbf{0.9715} & \textbf{0.9532}\\
    \midrule
        \multirow{3}{*}{VQEG HD3}& VMAF  & 0.9266  & 0.9238\\
        & AVQT & 0.9481  & 0.9417\\
        & VQT  & \textbf{0.9603} &  \textbf{0.9576} \\
    \midrule
        \multirow{3}{*}{Waterloo IVC 4k}& VMAF  & 0.7324  & 0.7325\\
        & AVQT & 0.7749  & 0.7738\\
        & VQT  & \textbf{0.7885} & \textbf{0.7821}\\
    \bottomrule
    \end{tabular}
    \label{tab:vmaf}
\end{table}

\subsection{Ablation Studies and Visualization}
To verify the rationality of the proposed modules, ablation studies are conducted in the following aspects. 
\paragraph{The rationale of the proposed STA module.} \label{ab1}
The rationale of STA is verified \textit{theoretically and experimentally}. \textit{Theoretically}, 
according to the JL lemma, the \textit{lower bound} of the error coefficient $\varepsilon$ after projection is measured by $ d >8 \ln (T) / \varepsilon^{2}$, where $T$ is the number of frames, and $d$ is the embedding size. In our setting of $T=96$ and $d=768$. So $\varepsilon$ can be calculated as 0.215, which means \textbf{more than 78.5\%} of temporal information or more is maintained by selected $\log T$ keyframes. \textit{Experimentally}, sufficient ablation studies also confirm the effectiveness of STA as shown in Tab.~\ref{tab:ab3} and \ref{tab:sta}. Simply adding the STA module to the baseline can bring consistent promotion on two VQA datasets (KoNViD-1k and Youtube-UGC) and one general video classification dataset (Kinetics-400).



\begin{table}[t]
    \centering
    \caption{Ablation study of individual modules in VQT, conducted in KoNViD-1k, Youtube-UGC, and Kinetics-400.}
    \setlength{\tabcolsep}{0.8mm}{
    \begin{tabular}{cc|cc|cc|c}
    \toprule
    \multicolumn{2}{c|}{Modules} & \multicolumn{2}{c|}{KoNViD-1k} & \multicolumn{2}{c|}{YoutuUGC} & K400\\
    \midrule
    STA & MPTN  & PLCC $\uparrow$ & SROCC $\uparrow$ & PLCC $\uparrow$ & SROCC $\uparrow$ & Top-1 Acc $\uparrow$ \\
    \midrule
    $\checkmark$ & $\checkmark$ & \textbf{0.8684} & \textbf{0.8582} & \textbf{0.8514} & \textbf{0.8357} & \textbf{80.3} \\ 
    \midrule
    $\checkmark$ & $\times$ & 0.8530 & 0.8510 & 0.8429 & 0.8280 & 79.0 \\ 
    $\times$ & $\times$ & 0.8293 & 0.8342  & 0.8279 & 0.8133 & 78.0\\ 
    \bottomrule
    \end{tabular}}
    \label{tab:ab3}
\end{table}


\begin{table}[t]
    \centering
    \caption{Ablation study on the type of reduction and the number of frames used in the STA module. Experiments are conducted in the KoNViD-1k.}
    \begin{tabular}{c|c|cc}
    \toprule
        Sampling Strategy & Frames & PLCC$\uparrow$ & SROCC$\uparrow$ \\
    \midrule
        our KL-based & $\log T$ & \textbf{0.8684} & \textbf{0.8582} \\
    \midrule
        our KL-based & $0.5\log T$ & 0.8320 & 0.8309 \\
        our KL-based & $2\log T$ & 0.8691 & 0.8575 \\
    \midrule
        Random\cite{DBLP:conf/iclr/ChoromanskiLDSG21}      & $\log T$ &  0.8067 & 0.7798 \\
        Linear\cite{DBLP:journals/corr/abs-2006-04768}      & $\log T$ & 0.8142 & 0.8091 \\
        Conv\cite{DBLP:journals/corr/abs-2102-12122}        & $\log T$ & 0.8140 & 0.8090 \\
        Clustering  & $\log T$ & 0.8379 & 0.8203 \\
    \bottomrule
    \end{tabular}
    \label{tab:sta}
\end{table}

\paragraph{Reduction types in the STA module.}
We conducted an evaluation of various linear reduction methods, including 
\textit{random}\cite{DBLP:conf/iclr/ChoromanskiLDSG21} ($\log T$ frames are randomly selected from $T$ frames), 
\textit{linear reduction}\cite{DBLP:journals/corr/abs-2006-04768} (features are transformed using a matrix of $\log T \times T$, resulting in the representation of $\log T$ frames), 
\textit{conv reduction}\cite{DBLP:journals/corr/abs-2102-12122} (features are transformed using a Conv/BN/ReLU module, which reduces the channel from $T$ to $\log T$), 
\textit{clustering} (features are clustered into $\log T$ centers according to cosine similarity), and \textit{STA}. The results, presented in Tab.~\ref{tab:sta}, demonstrate that the \textit{STA} reduction method has the most significant positive impact on performance, effectively removing redundant information. Additionally, we found that the $\log{T}$ setting was the optimal sparsity setting through experimentation involving increasing and decreasing the number of frames used to represent keyframes.



\begin{figure*}[t]
\begin{center}
\includegraphics[width=0.93\linewidth]{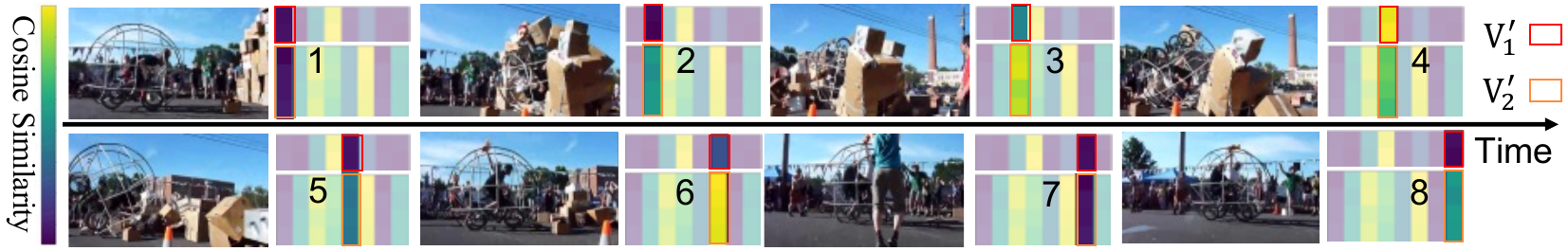}
\end{center}
  \caption{Visualization results of the temporal attention maps generated by an MPTN consisting of two STA modules (No.5956265529 in the KoNViD-1k). By the comparison between $\mathbf{V}_1^{\prime}$ and $\mathbf{V}_2^{\prime}$, these two STA modules concentrate on different distortions. The STA with fewer frames pays much attention to compression distortions that can be detected by spatial information, and the STA with more frames focuses on camera movement that can be detected by temporal information.
  }
\label{fig:learned_vis}
\end{figure*}

\paragraph{Different combinations of proposed modules.}
We conduct ablation experiments for different proposed modules, including STA and MPTN. Results are given in Tab.~\ref{tab:ab3}. The best performance is obtained by combining them. We also verify individual modules on Kinetics-400, where a combination of modules improves 1.3\% on Top-1 accuracy which proves the ability of generalization on general classification tasks.
For better understanding, we plot the visualization of the learned temporal attention maps in Fig. \ref{fig:learned_vis}. The visualization shows that different STA modules in the MPTN pay attention to different frames in a clip. It means that the proposed MPTN can capture different distortions simultaneously.

\begin{figure}[t]
\begin{center}
\includegraphics[width=\linewidth]{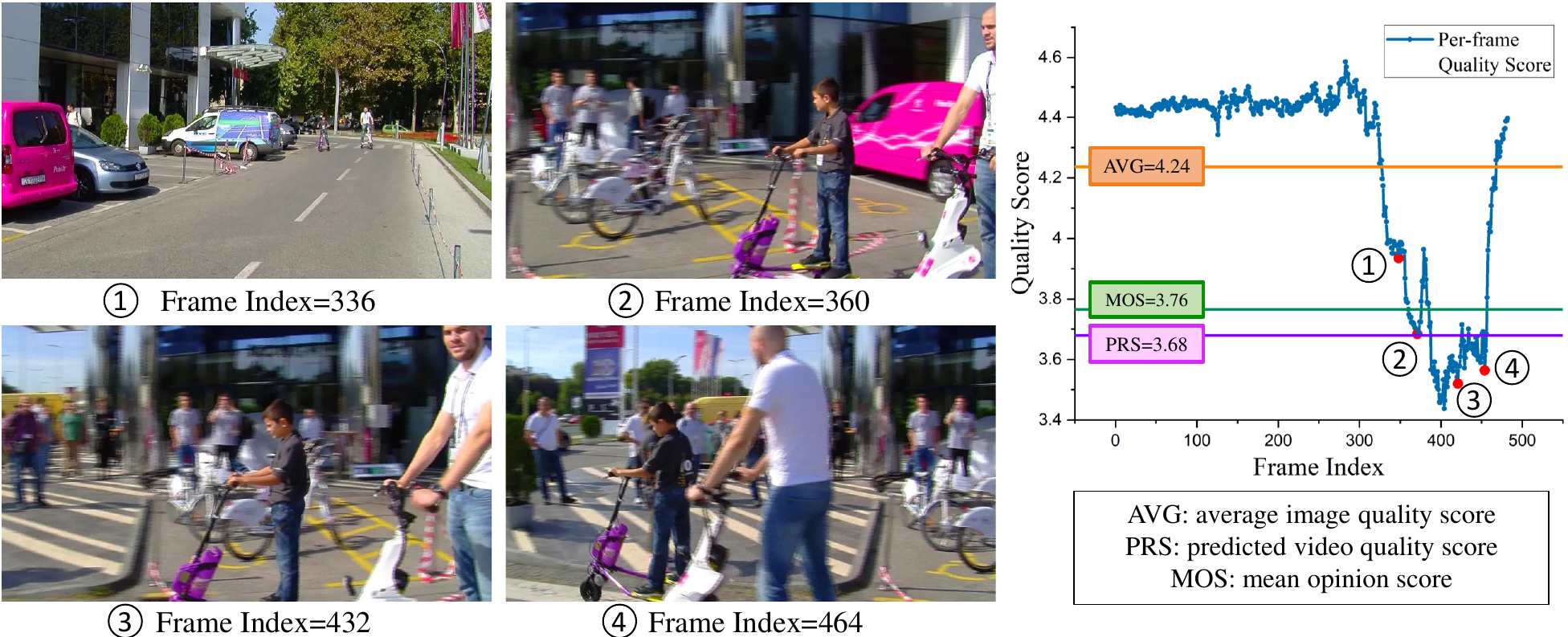}
\end{center}
\caption{Visualization results of the No.TelevisionClip-0604 video sampled from Youtube-UGC. The mixture of interlace and motion blur deteriorates the video quality. Compared to the average image quality score, VQT focuses more on the clips containing distortions, as shown by 4 sampled frames.
}
\label{fig:ytugcvis}
\end{figure}

\begin{figure}[t]
\begin{center}
\includegraphics[width=\linewidth]{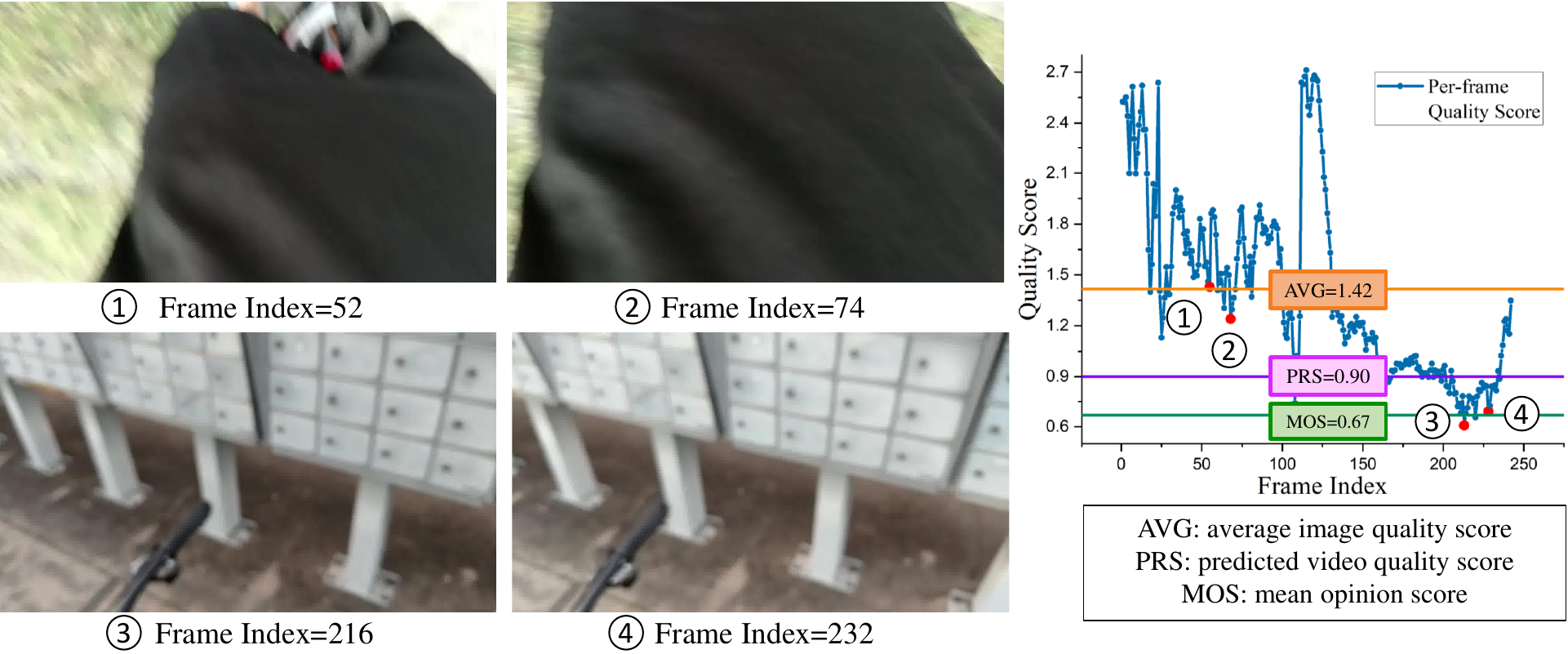}
\end{center}
\caption{Visualization results of the No.B174 video sampled from LIVE-VQC. Our analysis indicates that the primary factors contributing to the degradation are motion blur (exhibited in the 1st and 2nd frames) and out-of-focus (exhibited in the 3rd and 4th frames).}
\label{fig:livqvqcvis}
\end{figure}

\begin{figure}[t]
\begin{center}
\includegraphics[width=\linewidth]{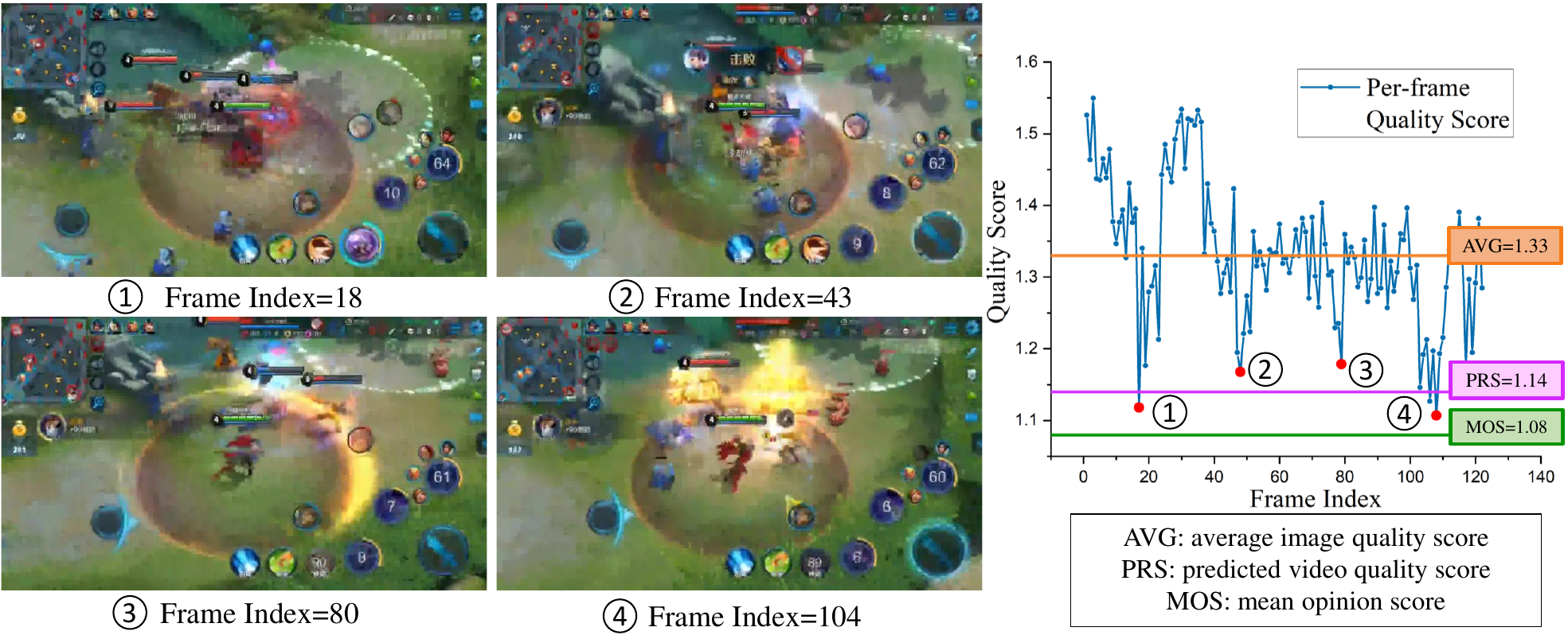}
\end{center}
\caption{Visualization results of the No.0095\_47 video sampled from ICME. The video quality is mainly decided by blocking artifacts, which appear in the fighting scenarios.
}
\label{fig:icmevis}
\end{figure}

\paragraph{Visualization} 
To further validate the effectiveness of VQT, we visualize the frames with the highest response and analyze the corresponding quality scores for each individual frame. As shown in Fig.~\ref{fig:ytugcvis}, \ref{fig:livqvqcvis} and \ref{fig:icmevis}, VQT is capable of effectively perceiving co-existing low-quality features in videos, such as interlace, motion blur, out-of-defocus and blocking artifacts. Furthermore, compared to averaging the predicted results for all frames, VQT-based temporal processing yields prediction results that are closer to the labeled MOS values.


\paragraph{Efficiency comparison.}
Computation Efficiency is compared with SoTA methods under 1080P/30FPS/30s videos, as shown in Tab.~\ref{tab:speed}, including academic algorithms and industrial algorithms. Due to commercial confidentiality, we cannot obtain open-source model information (\eg, FLOPs) from Netflix/Apple. For a fair comparison, in GPU, VQT shows higher efficiency than image-based (MDTSVAF, STDAM) and video-based SoTA methods (StarVQA, BVQA). That $0.5$s inference time can fulfill the real-time monitoring of the service-side quality variation. Further, speed-up can be investigated in future work (\eg, knowledge distillation and quantization). 

\begin{table}[t]
    \centering
    \caption{Comparison of inference cost with academic and industrial methods. VQT shows high efficiency.}
    \begin{tabular}{c|cc|cc}
    \toprule
        Type & TFLOPs & Device & Time & Speed Ratio\\
    \midrule
        VMAF & - & CPU & 9.85s & 1.0$\times$ \\
        AVQT & - & CPU & 4.61s & 2.1$\times$ \\
    \midrule
        MDTSVFA & 231 & GPU & 7.07s & 1.4$\times$  \\
        StarVQA & 197 & GPU & 0.57s & 17.3$\times$ \\
        BVQA    & 89  & GPU & 0.51s & 19.3$\times$ \\
        STDAM   & 106 & GPU & 2.12s & 4.6$\times$  \\
    \bottomrule
        VQT     & 154 & GPU & 0.50s & 19.7$\times$ \\
    \hline
    \end{tabular}
    \label{tab:speed}
\end{table}

\subsection{Generalization in Video Classification}
To further evaluate the generalization ability of VQT to other general semantic task, the performance on the video classification dataset of Kinetics-400\cite{DBLP:conf/cvpr/CarreiraZ17} with SoTA video classification models is evaludated, including R(2+1)D \cite{DBLP:conf/cvpr/TranWTRLP18}, I3D \cite{DBLP:conf/cvpr/CarreiraZ17}, I3D+NL \cite{DBLP:conf/cvpr/0004GGH18}, ip-CSN-152 \cite{DBLP:conf/iccv/TranWFT19}, SlowFast \cite{DBLP:conf/iccv/Feichtenhofer0M19}, TimeSformer \cite{DBLP:conf/icml/BertasiusWT21} and Video Swin Transformer \cite{DBLP:journals/corr/abs-2106-13230}.
We follow the default training setting in MMAction2 for a fair comparison.
As shown in Tab.~\ref{tab:k400}, VQT achieves a Top-1 accuracy by 80.3\%, outperforming CNN models \cite{DBLP:conf/iccv/TranWFT19,DBLP:conf/cvpr/TranWTRLP18,DBLP:conf/eccv/XieSHTM18} and very recent Transformer-based models, \eg TimeSformer\cite{DBLP:conf/icml/BertasiusWT21} and Video Swin Transformer\cite{DBLP:journals/corr/abs-2106-13230}. 
This proves the effectiveness and generalization ability of VQT in the general video classification task. We hope this VQT module can achieve more satisfactory performance when used in more general computer vision tasks.

\begin{table}[t]
    \centering
    \caption{Classification results on the K-400 validation set. The computational FLOPs and Top-k accuracy are reported.}
    \begin{tabular}{c|c|cc|c}
    \toprule
        Method & Backbone &  Top-1 & Top-5 & TFLOPs \\
    \midrule
        R(2+1)D & ResNet34 & 72.0 & 90.0 & 75 \\
        I3D     & ResNet50 & 72.1 & 90.3 & 108 \\ 
        I3D+NL  & ResNet101 & 77.7 & 93.3 & 359 \\
        ip-CSN-152 & Res152 & 77.8 & 92.8 & 109 \\
        SlowFast   & ResNet101 & 79.8 & 93.9 & 234 \\
        TimeSformer   & ViT-B & 78.0 & 93.7 & 197 \\
        Video Swin   & Swin-T & 78.8 & 93.6 & 88 \\
    \midrule
        VQT & ViT-B &\textbf{ 80.3} & \textbf{94.5} &  \textbf{154} \\
    \bottomrule
    \end{tabular}
    \label{tab:k400}
\end{table}

\section{Conclusion and Future Work}

This paper proposed VQT to address two underestimated challenges faced by VQA. To tackle the first challenge that the perceptual quality of videos is largely determined by deteriorated keyframes, we propose the STA module, which performs sparse sampling by analyzing the correlation between frames, resulting in efficient computation of attention. To address the second challenge that various types of distortions co-exist in a video, we propose the MPTN capture co-existing distortions by stacking multiple STA modules with different degrees of sparsity. Our proposed method is extensively evaluated on three widely-used NR-VQA datasets. Additionally, VQT outperforms widely-adopted industrial algorithms of VMAF and AVQT on four FR-VQA datasets. Extensive ablation studies and visual analysis further validate the effectiveness of each component of VQT. We also observe good generalization ability when transferring to the video classification task. We hope that VQT can serve as a new baseline for VQA tasks.

Regarding the selection of keyframes, the STA module currently relies on predefined hyperparameters. In the future, it would be possible to propose methods for adaptive keyframe selection, which can determine the number of keyframes based on prior information or employ a learning-based approach to identify keyframes that exhibit distortion-related features.



\clearpage
\bibliographystyle{ACM-Reference-Format}
\bibliography{paper}


\begin{thebibliography}{70}


\ifx \showCODEN    \undefined \def \showCODEN     #1{\unskip}     \fi
\ifx \showDOI      \undefined \def \showDOI       #1{#1}\fi
\ifx \showISBNx    \undefined \def \showISBNx     #1{\unskip}     \fi
\ifx \showISBNxiii \undefined \def \showISBNxiii  #1{\unskip}     \fi
\ifx \showISSN     \undefined \def \showISSN      #1{\unskip}     \fi
\ifx \showLCCN     \undefined \def \showLCCN      #1{\unskip}     \fi
\ifx \shownote     \undefined \def \shownote      #1{#1}          \fi
\ifx \showarticletitle \undefined \def \showarticletitle #1{#1}   \fi
\ifx \showURL      \undefined \def \showURL       {\relax}        \fi
\providecommand\bibfield[2]{#2}
\providecommand\bibinfo[2]{#2}
\providecommand\natexlab[1]{#1}
\providecommand\showeprint[2][]{arXiv:#2}

\bibitem[Amer and Dubois(2005)]%
        {DBLP:journals/tcsv/AmerD05}
\bibfield{author}{\bibinfo{person}{Aishy Amer} {and} \bibinfo{person}{Eric
  Dubois}.} \bibinfo{year}{2005}\natexlab{}.
\newblock \showarticletitle{Fast and reliable structure-oriented video noise
  estimation}.
\newblock \bibinfo{journal}{\emph{{IEEE} TCSVT}} \bibinfo{volume}{15},
  \bibinfo{number}{1} (\bibinfo{year}{2005}), \bibinfo{pages}{113--118}.
\newblock


\bibitem[Arnab et~al\mbox{.}(2021)]%
        {DBLP:conf/iccv/Arnab0H0LS21}
\bibfield{author}{\bibinfo{person}{Anurag Arnab}, \bibinfo{person}{Mostafa
  Dehghani}, \bibinfo{person}{Georg Heigold}, \bibinfo{person}{Chen Sun},
  \bibinfo{person}{Mario Lucic}, {and} \bibinfo{person}{Cordelia Schmid}.}
  \bibinfo{year}{2021}\natexlab{}.
\newblock \showarticletitle{ViViT: {A} Video Vision Transformer}. In
  \bibinfo{booktitle}{\emph{{ICCV}}}. \bibinfo{publisher}{{IEEE}},
  \bibinfo{pages}{6816--6826}.
\newblock


\bibitem[Bertasius et~al\mbox{.}(2021)]%
        {DBLP:conf/icml/BertasiusWT21}
\bibfield{author}{\bibinfo{person}{Gedas Bertasius}, \bibinfo{person}{Heng
  Wang}, {and} \bibinfo{person}{Lorenzo Torresani}.}
  \bibinfo{year}{2021}\natexlab{}.
\newblock \showarticletitle{Is Space-Time Attention All You Need for Video
  Understanding?}. In \bibinfo{booktitle}{\emph{{ICML}}},
  Vol.~\bibinfo{volume}{139}. \bibinfo{publisher}{{PMLR}},
  \bibinfo{pages}{813--824}.
\newblock


\bibitem[Bosse et~al\mbox{.}(2016)]%
        {DBLP:conf/icip/BosseMWS16}
\bibfield{author}{\bibinfo{person}{Sebastian Bosse}, \bibinfo{person}{Dominique
  Maniry}, \bibinfo{person}{Thomas Wiegand}, {and} \bibinfo{person}{Wojciech
  Samek}.} \bibinfo{year}{2016}\natexlab{}.
\newblock \showarticletitle{A deep neural network for image quality
  assessment}. In \bibinfo{booktitle}{\emph{{ICIP}}}.
  \bibinfo{publisher}{{IEEE}}, \bibinfo{pages}{3773--3777}.
\newblock


\bibitem[Carreira and Zisserman(2017)]%
        {DBLP:conf/cvpr/CarreiraZ17}
\bibfield{author}{\bibinfo{person}{Jo{\~{a}}o Carreira} {and}
  \bibinfo{person}{Andrew Zisserman}.} \bibinfo{year}{2017}\natexlab{}.
\newblock \showarticletitle{Quo Vadis, Action Recognition? {A} New Model and
  the Kinetics Dataset}. In \bibinfo{booktitle}{\emph{{CVPR}}}.
  \bibinfo{publisher}{{IEEE} Computer Society}, \bibinfo{pages}{4724--4733}.
\newblock


\bibitem[Chadha and Andreopoulos(2021)]%
        {DBLP:conf/cvpr/ChadhaA21}
\bibfield{author}{\bibinfo{person}{Aaron Chadha} {and} \bibinfo{person}{Yiannis
  Andreopoulos}.} \bibinfo{year}{2021}\natexlab{}.
\newblock \showarticletitle{Deep Perceptual Preprocessing for Video Coding}. In
  \bibinfo{booktitle}{\emph{{CVPR}}}. \bibinfo{publisher}{Computer Vision
  Foundation / {IEEE}}, \bibinfo{pages}{14852--14861}.
\newblock


\bibitem[Choromanski et~al\mbox{.}(2021)]%
        {DBLP:conf/iclr/ChoromanskiLDSG21}
\bibfield{author}{\bibinfo{person}{Krzysztof~Marcin Choromanski},
  \bibinfo{person}{Valerii Likhosherstov}, \bibinfo{person}{David Dohan},
  \bibinfo{person}{Xingyou Song}, \bibinfo{person}{Andreea Gane},
  \bibinfo{person}{Tam{\'{a}}s Sarl{\'{o}}s}, \bibinfo{person}{Peter Hawkins},
  \bibinfo{person}{Jared~Quincy Davis}, \bibinfo{person}{Afroz Mohiuddin},
  \bibinfo{person}{Lukasz Kaiser}, \bibinfo{person}{David~Benjamin Belanger},
  \bibinfo{person}{Lucy~J. Colwell}, {and} \bibinfo{person}{Adrian Weller}.}
  \bibinfo{year}{2021}\natexlab{}.
\newblock \showarticletitle{Rethinking Attention with Performers}. In
  \bibinfo{booktitle}{\emph{{ICLR}}}.
\newblock


\bibitem[Contributors(2020)]%
        {2020mmaction2}
\bibfield{author}{\bibinfo{person}{MMAction2 Contributors}.}
  \bibinfo{year}{2020}\natexlab{}.
\newblock \bibinfo{title}{OpenMMLab's Next Generation Video Understanding
  Toolbox and Benchmark}.
\newblock
  \bibinfo{howpublished}{\url{https://github.com/open-mmlab/mmaction2}}.
\newblock


\bibitem[Deng et~al\mbox{.}(2009)]%
        {imagenet}
\bibfield{author}{\bibinfo{person}{Jia Deng}, \bibinfo{person}{Wei Dong},
  \bibinfo{person}{Richard Socher}, \bibinfo{person}{Li{-}Jia Li},
  \bibinfo{person}{Kai Li}, {and} \bibinfo{person}{Li Fei{-}Fei}.}
  \bibinfo{year}{2009}\natexlab{}.
\newblock \showarticletitle{ImageNet: {A} large-scale hierarchical image
  database}. In \bibinfo{booktitle}{\emph{{CVPR}}}. \bibinfo{publisher}{{IEEE}
  Computer Society}, \bibinfo{pages}{248--255}.
\newblock


\bibitem[Dosovitskiy et~al\mbox{.}(2021)]%
        {DBLP:conf/iclr/DosovitskiyB0WZ21}
\bibfield{author}{\bibinfo{person}{Alexey Dosovitskiy}, \bibinfo{person}{Lucas
  Beyer}, \bibinfo{person}{Alexander Kolesnikov}, \bibinfo{person}{Dirk
  Weissenborn}, \bibinfo{person}{Xiaohua Zhai}, \bibinfo{person}{Thomas
  Unterthiner}, \bibinfo{person}{Mostafa Dehghani}, \bibinfo{person}{Matthias
  Minderer}, \bibinfo{person}{Georg Heigold}, \bibinfo{person}{Sylvain Gelly},
  \bibinfo{person}{Jakob Uszkoreit}, {and} \bibinfo{person}{Neil Houlsby}.}
  \bibinfo{year}{2021}\natexlab{}.
\newblock \showarticletitle{An Image is Worth 16x16 Words: Transformers for
  Image Recognition at Scale}. In \bibinfo{booktitle}{\emph{{ICLR}}}.
\newblock


\bibitem[Ebenezer et~al\mbox{.}(2020)]%
        {DBLP:conf/mmsp/EbenezerSWWB20}
\bibfield{author}{\bibinfo{person}{Joshua~Peter Ebenezer},
  \bibinfo{person}{Zaixi Shang}, \bibinfo{person}{Yongjun Wu},
  \bibinfo{person}{Hai Wei}, {and} \bibinfo{person}{Alan~C. Bovik}.}
  \bibinfo{year}{2020}\natexlab{}.
\newblock \showarticletitle{No-Reference Video Quality Assessment Using
  Space-Time Chips}. In \bibinfo{booktitle}{\emph{{MMSP}}}.
  \bibinfo{publisher}{{IEEE}}, \bibinfo{pages}{1--6}.
\newblock


\bibitem[Esfandarani and Milanfar(2018)]%
        {DBLP:conf/iccp/TalebiM18}
\bibfield{author}{\bibinfo{person}{Hossein~Talebi Esfandarani} {and}
  \bibinfo{person}{Peyman Milanfar}.} \bibinfo{year}{2018}\natexlab{}.
\newblock \showarticletitle{Learned perceptual image enhancement}. In
  \bibinfo{booktitle}{\emph{{ICCP}}}. \bibinfo{pages}{1--13}.
\newblock


\bibitem[Fan et~al\mbox{.}(2021)]%
        {Fan_2021_ICCV}
\bibfield{author}{\bibinfo{person}{Haoqi Fan}, \bibinfo{person}{Bo Xiong},
  \bibinfo{person}{Karttikeya Mangalam}, \bibinfo{person}{Yanghao Li},
  \bibinfo{person}{Zhicheng Yan}, \bibinfo{person}{Jitendra Malik}, {and}
  \bibinfo{person}{Christoph Feichtenhofer}.} \bibinfo{year}{2021}\natexlab{}.
\newblock \showarticletitle{Multiscale Vision Transformers}. In
  \bibinfo{booktitle}{\emph{ICCV}}. \bibinfo{pages}{6824--6835}.
\newblock


\bibitem[Fang et~al\mbox{.}(2020)]%
        {DBLP:conf/cvpr/FangZZMW20}
\bibfield{author}{\bibinfo{person}{Yuming Fang}, \bibinfo{person}{Hanwei Zhu},
  \bibinfo{person}{Yan Zeng}, \bibinfo{person}{Kede Ma}, {and}
  \bibinfo{person}{Zhou Wang}.} \bibinfo{year}{2020}\natexlab{}.
\newblock \showarticletitle{Perceptual Quality Assessment of Smartphone
  Photography}. In \bibinfo{booktitle}{\emph{{CVPR}}}.
  \bibinfo{publisher}{Computer Vision Foundation / {IEEE}},
  \bibinfo{pages}{3674--3683}.
\newblock


\bibitem[Feichtenhofer et~al\mbox{.}(2019)]%
        {DBLP:conf/iccv/Feichtenhofer0M19}
\bibfield{author}{\bibinfo{person}{Christoph Feichtenhofer},
  \bibinfo{person}{Haoqi Fan}, \bibinfo{person}{Jitendra Malik}, {and}
  \bibinfo{person}{Kaiming He}.} \bibinfo{year}{2019}\natexlab{}.
\newblock \showarticletitle{SlowFast Networks for Video Recognition}. In
  \bibinfo{booktitle}{\emph{{ICCV}}}. \bibinfo{pages}{6201--6210}.
\newblock


\bibitem[Gao et~al\mbox{.}(2017)]%
        {DBLP:journals/ijon/GaoWLTYZ17}
\bibfield{author}{\bibinfo{person}{Fei Gao}, \bibinfo{person}{Yi Wang},
  \bibinfo{person}{Panpeng Li}, \bibinfo{person}{Min Tan}, \bibinfo{person}{Jun
  Yu}, {and} \bibinfo{person}{Yani Zhu}.} \bibinfo{year}{2017}\natexlab{}.
\newblock \showarticletitle{DeepSim: Deep similarity for image quality
  assessment}.
\newblock \bibinfo{journal}{\emph{Neurocomputing}}  \bibinfo{volume}{257}
  (\bibinfo{year}{2017}), \bibinfo{pages}{104--114}.
\newblock


\bibitem[Ghadiyaram et~al\mbox{.}(2019)]%
        {DBLP:journals/tcsv/GhadiyaramPB19}
\bibfield{author}{\bibinfo{person}{Deepti Ghadiyaram}, \bibinfo{person}{Janice
  Pan}, {and} \bibinfo{person}{Alan~C. Bovik}.}
  \bibinfo{year}{2019}\natexlab{}.
\newblock \showarticletitle{A Subjective and Objective Study of Stalling Events
  in Mobile Streaming Videos}.
\newblock \bibinfo{journal}{\emph{{IEEE} Trans. Circuits Syst. Video Technol.}}
  \bibinfo{volume}{29}, \bibinfo{number}{1} (\bibinfo{year}{2019}),
  \bibinfo{pages}{183--197}.
\newblock


\bibitem[Gu et~al\mbox{.}(2019)]%
        {DBLP:conf/aaai/GuMDXP19}
\bibfield{author}{\bibinfo{person}{Jie Gu}, \bibinfo{person}{Gaofeng Meng},
  \bibinfo{person}{Cheng Da}, \bibinfo{person}{Shiming Xiang}, {and}
  \bibinfo{person}{Chunhong Pan}.} \bibinfo{year}{2019}\natexlab{}.
\newblock \showarticletitle{No-Reference Image Quality Assessment with
  Reinforcement Recursive List-Wise Ranking}. In
  \bibinfo{booktitle}{\emph{{AAAI}}}. \bibinfo{publisher}{{AAAI} Press},
  \bibinfo{pages}{8336--8343}.
\newblock


\bibitem[Gu et~al\mbox{.}(2009)]%
        {DBLP:journals/tog/GuRBN09}
\bibfield{author}{\bibinfo{person}{Jinwei Gu}, \bibinfo{person}{Ravi
  Ramamoorthi}, \bibinfo{person}{Peter~N. Belhumeur}, {and}
  \bibinfo{person}{Shree~K. Nayar}.} \bibinfo{year}{2009}\natexlab{}.
\newblock \showarticletitle{Removing image artifacts due to dirty camera lenses
  and thin occluders}.
\newblock \bibinfo{journal}{\emph{{ACM} Trans. Graph.}} \bibinfo{volume}{28},
  \bibinfo{number}{5} (\bibinfo{year}{2009}), \bibinfo{pages}{144}.
\newblock


\bibitem[Hosu et~al\mbox{.}(2017)]%
        {DBLP:conf/qomex/HosuHJLMSLS17}
\bibfield{author}{\bibinfo{person}{Vlad Hosu}, \bibinfo{person}{Franz Hahn},
  \bibinfo{person}{Mohsen Jenadeleh}, \bibinfo{person}{Hanhe Lin},
  \bibinfo{person}{Hui Men}, \bibinfo{person}{Tam{\'{a}}s Szir{\'{a}}nyi},
  \bibinfo{person}{Shujun Li}, {and} \bibinfo{person}{Dietmar Saupe}.}
  \bibinfo{year}{2017}\natexlab{}.
\newblock \showarticletitle{The Konstanz natural video database (KoNViD-1k)}.
  In \bibinfo{booktitle}{\emph{QoMEX}}. \bibinfo{publisher}{{IEEE}},
  \bibinfo{pages}{1--6}.
\newblock


\bibitem[Johnson et~al\mbox{.}(2016)]%
        {DBLP:conf/eccv/JohnsonAF16}
\bibfield{author}{\bibinfo{person}{Justin Johnson}, \bibinfo{person}{Alexandre
  Alahi}, {and} \bibinfo{person}{Li Fei{-}Fei}.}
  \bibinfo{year}{2016}\natexlab{}.
\newblock \showarticletitle{Perceptual Losses for Real-Time Style Transfer and
  Super-Resolution}. In \bibinfo{booktitle}{\emph{{ECCV}}}.
  \bibinfo{pages}{694--711}.
\newblock


\bibitem[Johnson and Lindenstrauss(1984)]%
        {jllemma1984}
\bibfield{author}{\bibinfo{person}{William Johnson} {and}
  \bibinfo{person}{Joram Lindenstrauss}.} \bibinfo{year}{1984}\natexlab{}.
\newblock \showarticletitle{Extensions of Lipschitz maps into a Hilbert space}.
\newblock \bibinfo{journal}{\emph{Contemp. Math.}}  \bibinfo{volume}{26}
  (\bibinfo{year}{1984}).
\newblock


\bibitem[Kang et~al\mbox{.}(2014)]%
        {DBLP:conf/cvpr/KangYLD14}
\bibfield{author}{\bibinfo{person}{Le Kang}, \bibinfo{person}{Peng Ye},
  \bibinfo{person}{Yi Li}, {and} \bibinfo{person}{David~S. Doermann}.}
  \bibinfo{year}{2014}\natexlab{}.
\newblock \showarticletitle{Convolutional Neural Networks for No-Reference
  Image Quality Assessment}. In \bibinfo{booktitle}{\emph{{CVPR}}}.
  \bibinfo{publisher}{{IEEE} Computer Society}, \bibinfo{pages}{1733--1740}.
\newblock


\bibitem[Kay et~al\mbox{.}(2017)]%
        {kinetics}
\bibfield{author}{\bibinfo{person}{Will Kay}, \bibinfo{person}{Jo{\~{a}}o
  Carreira}, \bibinfo{person}{Karen Simonyan}, \bibinfo{person}{Brian Zhang},
  \bibinfo{person}{Chloe Hillier}, \bibinfo{person}{Sudheendra
  Vijayanarasimhan}, \bibinfo{person}{Fabio Viola}, \bibinfo{person}{Tim
  Green}, \bibinfo{person}{Trevor Back}, \bibinfo{person}{Paul Natsev},
  \bibinfo{person}{Mustafa Suleyman}, {and} \bibinfo{person}{Andrew
  Zisserman}.} \bibinfo{year}{2017}\natexlab{}.
\newblock \showarticletitle{The Kinetics Human Action Video Dataset}.
\newblock \bibinfo{journal}{\emph{CoRR}}  \bibinfo{volume}{abs/1705.06950}
  (\bibinfo{year}{2017}).
\newblock


\bibitem[Ke et~al\mbox{.}(2021)]%
        {DBLP:conf/iccv/KeWWMY21}
\bibfield{author}{\bibinfo{person}{Junjie Ke}, \bibinfo{person}{Qifei Wang},
  \bibinfo{person}{Yilin Wang}, \bibinfo{person}{Peyman Milanfar}, {and}
  \bibinfo{person}{Feng Yang}.} \bibinfo{year}{2021}\natexlab{}.
\newblock \showarticletitle{{MUSIQ:} Multi-scale Image Quality Transformer}. In
  \bibinfo{booktitle}{\emph{{ICCV}}}. \bibinfo{pages}{5128--5137}.
\newblock


\bibitem[Kim and Lee(2017)]%
        {DBLP:conf/cvpr/Kim017}
\bibfield{author}{\bibinfo{person}{Jongyoo Kim} {and} \bibinfo{person}{Sanghoon
  Lee}.} \bibinfo{year}{2017}\natexlab{}.
\newblock \showarticletitle{Deep Learning of Human Visual Sensitivity in Image
  Quality Assessment Framework}. In \bibinfo{booktitle}{\emph{{CVPR}}}.
  \bibinfo{pages}{1969--1977}.
\newblock


\bibitem[Kim et~al\mbox{.}(2018)]%
        {DBLP:conf/eccv/KimKAKL18}
\bibfield{author}{\bibinfo{person}{Woojae Kim}, \bibinfo{person}{Jongyoo Kim},
  \bibinfo{person}{Sewoong Ahn}, \bibinfo{person}{Jinwoo Kim}, {and}
  \bibinfo{person}{Sanghoon Lee}.} \bibinfo{year}{2018}\natexlab{}.
\newblock \showarticletitle{Deep Video Quality Assessor: From Spatio-Temporal
  Visual Sensitivity to a Convolutional Neural Aggregation Network}. In
  \bibinfo{booktitle}{\emph{ECCV}}, Vol.~\bibinfo{volume}{11205}.
  \bibinfo{pages}{224--241}.
\newblock


\bibitem[Li et~al\mbox{.}(2021b)]%
        {DBLP:journals/corr/abs-2108-08505}
\bibfield{author}{\bibinfo{person}{Bowen Li}, \bibinfo{person}{Weixia Zhang},
  \bibinfo{person}{Meng Tian}, \bibinfo{person}{Guangtao Zhai}, {and}
  \bibinfo{person}{Xianpei Wang}.} \bibinfo{year}{2021}\natexlab{b}.
\newblock \showarticletitle{Blindly Assess Quality of In-the-Wild Videos via
  Quality-aware Pre-training and Motion Perception}.
\newblock \bibinfo{journal}{\emph{CoRR}}  \bibinfo{volume}{abs/2108.08505}
  (\bibinfo{year}{2021}).
\newblock


\bibitem[Li et~al\mbox{.}(2019b)]%
        {DBLP:conf/mm/LiJJ19}
\bibfield{author}{\bibinfo{person}{Dingquan Li}, \bibinfo{person}{Tingting
  Jiang}, {and} \bibinfo{person}{Ming Jiang}.}
  \bibinfo{year}{2019}\natexlab{b}.
\newblock \showarticletitle{Quality Assessment of In-the-Wild Videos}. In
  \bibinfo{booktitle}{\emph{{ACM} Multimedia}}. \bibinfo{publisher}{{ACM}},
  \bibinfo{pages}{2351--2359}.
\newblock


\bibitem[Li et~al\mbox{.}(2021a)]%
        {DBLP:journals/ijcv/LiJJ21}
\bibfield{author}{\bibinfo{person}{Dingquan Li}, \bibinfo{person}{Tingting
  Jiang}, {and} \bibinfo{person}{Ming Jiang}.}
  \bibinfo{year}{2021}\natexlab{a}.
\newblock \showarticletitle{Unified Quality Assessment of in-the-Wild Videos
  with Mixed Datasets Training}.
\newblock \bibinfo{journal}{\emph{IJCV}}  \bibinfo{volume}{129}
  (\bibinfo{year}{2021}), \bibinfo{pages}{1238--1257}.
\newblock


\bibitem[Li et~al\mbox{.}(2016)]%
        {DBLP:conf/icdsp/LiPFY16}
\bibfield{author}{\bibinfo{person}{Yuming Li}, \bibinfo{person}{Lai{-}Man Po},
  \bibinfo{person}{Litong Feng}, {and} \bibinfo{person}{Fang Yuan}.}
  \bibinfo{year}{2016}\natexlab{}.
\newblock \showarticletitle{No-reference image quality assessment with deep
  convolutional neural networks}. In \bibinfo{booktitle}{\emph{{DSP}}}.
  \bibinfo{publisher}{{IEEE}}, \bibinfo{pages}{685--689}.
\newblock


\bibitem[Li et~al\mbox{.}(2019a)]%
        {waterloo}
\bibfield{author}{\bibinfo{person}{Zhuoran Li}, \bibinfo{person}{Zhengfang
  Duanmu}, \bibinfo{person}{Wentao Liu}, {and} \bibinfo{person}{Zhou Wang}.}
  \bibinfo{year}{2019}\natexlab{a}.
\newblock \showarticletitle{AVC, HEVC, VP9, {AVS2} or AV1? - {A} Comparative
  Study of State-of-the-Art Video Encoders on 4K Videos}. In
  \bibinfo{booktitle}{\emph{{ICIAR} {(1)}}} \emph{(\bibinfo{series}{Lecture
  Notes in Computer Science}, Vol.~\bibinfo{volume}{11662})}.
  \bibinfo{publisher}{Springer}, \bibinfo{pages}{162--173}.
\newblock


\bibitem[Liu et~al\mbox{.}(2014)]%
        {DBLP:conf/bmsb/LiuZZSGY14}
\bibfield{author}{\bibinfo{person}{Min Liu}, \bibinfo{person}{Guangtao Zhai},
  \bibinfo{person}{Zhenyu Zhang}, \bibinfo{person}{Yuntao Sun},
  \bibinfo{person}{Ke Gu}, {and} \bibinfo{person}{Xiaokang Yang}.}
  \bibinfo{year}{2014}\natexlab{}.
\newblock \showarticletitle{Blind image quality assessment for noise}. In
  \bibinfo{booktitle}{\emph{{BMSB}}}. \bibinfo{pages}{1--5}.
\newblock


\bibitem[Liu et~al\mbox{.}(2017)]%
        {DBLP:journals/jvcir/LiuGZLZG17}
\bibfield{author}{\bibinfo{person}{Yutao Liu}, \bibinfo{person}{Ke Gu},
  \bibinfo{person}{Guangtao Zhai}, \bibinfo{person}{Xianming Liu},
  \bibinfo{person}{Debin Zhao}, {and} \bibinfo{person}{Wen Gao}.}
  \bibinfo{year}{2017}\natexlab{}.
\newblock \showarticletitle{Quality assessment for real out-of-focus blurred
  images}.
\newblock \bibinfo{journal}{\emph{JVCIR}} (\bibinfo{year}{2017}).
\newblock


\bibitem[Liu et~al\mbox{.}(2021)]%
        {DBLP:journals/corr/abs-2106-13230}
\bibfield{author}{\bibinfo{person}{Ze Liu}, \bibinfo{person}{Jia Ning},
  \bibinfo{person}{Yue Cao}, \bibinfo{person}{Yixuan Wei},
  \bibinfo{person}{Zheng Zhang}, \bibinfo{person}{Stephen Lin}, {and}
  \bibinfo{person}{Han Hu}.} \bibinfo{year}{2021}\natexlab{}.
\newblock \showarticletitle{Video Swin Transformer}.
\newblock \bibinfo{journal}{\emph{CoRR}}  \bibinfo{volume}{abs/2106.13230}
  (\bibinfo{year}{2021}).
\newblock


\bibitem[Marziliano et~al\mbox{.}(2002)]%
        {DBLP:conf/icip/MarzilianoDWE02}
\bibfield{author}{\bibinfo{person}{Pina Marziliano},
  \bibinfo{person}{Fr{\'{e}}d{\'{e}}ric Dufaux}, \bibinfo{person}{Stefan
  Winkler}, {and} \bibinfo{person}{Touradj Ebrahimi}.}
  \bibinfo{year}{2002}\natexlab{}.
\newblock \showarticletitle{A no-reference perceptual blur metric}. In
  \bibinfo{booktitle}{\emph{ICIP}}. \bibinfo{pages}{57--60}.
\newblock


\bibitem[Min et~al\mbox{.}(2018)]%
        {DBLP:journals/sigpro/MinGZHY18}
\bibfield{author}{\bibinfo{person}{Xiongkuo Min}, \bibinfo{person}{Ke Gu},
  \bibinfo{person}{Guangtao Zhai}, \bibinfo{person}{Menghan Hu}, {and}
  \bibinfo{person}{Xiaokang Yang}.} \bibinfo{year}{2018}\natexlab{}.
\newblock \showarticletitle{Saliency-induced reduced-reference quality index
  for natural scene and screen content images}.
\newblock \bibinfo{journal}{\emph{SP}}  \bibinfo{volume}{145}
  (\bibinfo{year}{2018}), \bibinfo{pages}{127--136}.
\newblock


\bibitem[Mittal et~al\mbox{.}(2012)]%
        {DBLP:journals/tip/MittalMB12}
\bibfield{author}{\bibinfo{person}{Anish Mittal},
  \bibinfo{person}{Anush~Krishna Moorthy}, {and} \bibinfo{person}{Alan~Conrad
  Bovik}.} \bibinfo{year}{2012}\natexlab{}.
\newblock \showarticletitle{No-Reference Image Quality Assessment in the
  Spatial Domain}.
\newblock \bibinfo{journal}{\emph{{IEEE} Trans. Image Process.}}
  \bibinfo{volume}{21}, \bibinfo{number}{12} (\bibinfo{year}{2012}),
  \bibinfo{pages}{4695--4708}.
\newblock


\bibitem[Mittal et~al\mbox{.}(2016)]%
        {DBLP:journals/tip/MittalSB16}
\bibfield{author}{\bibinfo{person}{Anish Mittal}, \bibinfo{person}{Michele~A.
  Saad}, {and} \bibinfo{person}{Alan~C. Bovik}.}
  \bibinfo{year}{2016}\natexlab{}.
\newblock \showarticletitle{A Completely Blind Video Integrity Oracle}.
\newblock \bibinfo{journal}{\emph{{IEEE} TIP}} \bibinfo{volume}{25},
  \bibinfo{number}{1} (\bibinfo{year}{2016}), \bibinfo{pages}{289--300}.
\newblock


\bibitem[Mittal et~al\mbox{.}(2013)]%
        {DBLP:journals/spl/MittalSB13}
\bibfield{author}{\bibinfo{person}{Anish Mittal}, \bibinfo{person}{Rajiv
  Soundararajan}, {and} \bibinfo{person}{Alan~C. Bovik}.}
  \bibinfo{year}{2013}\natexlab{}.
\newblock \showarticletitle{Making a "Completely Blind" Image Quality
  Analyzer}.
\newblock \bibinfo{journal}{\emph{{IEEE} SPL}} \bibinfo{volume}{20},
  \bibinfo{number}{3} (\bibinfo{year}{2013}), \bibinfo{pages}{209--212}.
\newblock


\bibitem[Paszke et~al\mbox{.}(2019)]%
        {DBLP:conf/nips/PaszkeGMLBCKLGA19}
\bibfield{author}{\bibinfo{person}{Adam Paszke}, \bibinfo{person}{Sam Gross},
  \bibinfo{person}{Francisco Massa}, \bibinfo{person}{Adam Lerer},
  \bibinfo{person}{James Bradbury}, \bibinfo{person}{Gregory Chanan},
  \bibinfo{person}{Trevor Killeen}, \bibinfo{person}{Zeming Lin},
  \bibinfo{person}{Natalia Gimelshein}, \bibinfo{person}{Luca Antiga},
  \bibinfo{person}{Alban Desmaison}, \bibinfo{person}{Andreas K{\"{o}}pf},
  \bibinfo{person}{Edward~Z. Yang}, \bibinfo{person}{Zachary DeVito},
  \bibinfo{person}{Martin Raison}, \bibinfo{person}{Alykhan Tejani},
  \bibinfo{person}{Sasank Chilamkurthy}, \bibinfo{person}{Benoit Steiner},
  \bibinfo{person}{Lu Fang}, \bibinfo{person}{Junjie Bai}, {and}
  \bibinfo{person}{Soumith Chintala}.} \bibinfo{year}{2019}\natexlab{}.
\newblock \showarticletitle{PyTorch: An Imperative Style, High-Performance Deep
  Learning Library}. In \bibinfo{booktitle}{\emph{NeurIPS}}.
  \bibinfo{pages}{8024--8035}.
\newblock


\bibitem[Saad et~al\mbox{.}(2014)]%
        {DBLP:journals/tip/SaadBC14}
\bibfield{author}{\bibinfo{person}{Michele~A. Saad}, \bibinfo{person}{Alan~C.
  Bovik}, {and} \bibinfo{person}{Christophe Charrier}.}
  \bibinfo{year}{2014}\natexlab{}.
\newblock \showarticletitle{Blind Prediction of Natural Video Quality}.
\newblock \bibinfo{journal}{\emph{{IEEE} Trans. Image Process.}}
  \bibinfo{volume}{23}, \bibinfo{number}{3} (\bibinfo{year}{2014}),
  \bibinfo{pages}{1352--1365}.
\newblock


\bibitem[Sinno(2019)]%
        {DBLP:journals/tip/SinnoB19}
\bibfield{author}{\bibinfo{person}{Zeina Sinno}.}
  \bibinfo{year}{2019}\natexlab{}.
\newblock \showarticletitle{Large-Scale Study of Perceptual Video Quality}.
\newblock \bibinfo{journal}{\emph{{IEEE} TIP}} \bibinfo{volume}{28},
  \bibinfo{number}{2} (\bibinfo{year}{2019}), \bibinfo{pages}{612--627}.
\newblock


\bibitem[Telili et~al\mbox{.}(2022)]%
        {telili20222bivqa}
\bibfield{author}{\bibinfo{person}{Ahmed Telili}, \bibinfo{person}{Sid~Ahmed
  Fezza}, \bibinfo{person}{Wassim Hamidouche}, {and} \bibinfo{person}{Hanene~FZ
  Meftah}.} \bibinfo{year}{2022}\natexlab{}.
\newblock \showarticletitle{2BiVQA: Double Bi-LSTM based Video Quality
  Assessment of UGC Videos}.
\newblock \bibinfo{journal}{\emph{arXiv preprint arXiv:2208.14774}}
  (\bibinfo{year}{2022}).
\newblock


\bibitem[Thomee et~al\mbox{.}(2016)]%
        {DBLP:journals/cacm/ThomeeSFENPBL16}
\bibfield{author}{\bibinfo{person}{Bart Thomee}, \bibinfo{person}{David~A.
  Shamma}, \bibinfo{person}{Gerald Friedland}, \bibinfo{person}{Benjamin
  Elizalde}, \bibinfo{person}{Karl Ni}, \bibinfo{person}{Douglas Poland},
  \bibinfo{person}{Damian Borth}, {and} \bibinfo{person}{Li{-}Jia Li}.}
  \bibinfo{year}{2016}\natexlab{}.
\newblock \showarticletitle{{YFCC100M:} the new data in multimedia research}.
\newblock \bibinfo{journal}{\emph{Commun. {ACM}}} \bibinfo{volume}{59},
  \bibinfo{number}{2} (\bibinfo{year}{2016}), \bibinfo{pages}{64--73}.
\newblock


\bibitem[Tran et~al\mbox{.}(2019)]%
        {DBLP:conf/iccv/TranWFT19}
\bibfield{author}{\bibinfo{person}{Du Tran}, \bibinfo{person}{Heng Wang},
  \bibinfo{person}{Matt Feiszli}, {and} \bibinfo{person}{Lorenzo Torresani}.}
  \bibinfo{year}{2019}\natexlab{}.
\newblock \showarticletitle{Video Classification With Channel-Separated
  Convolutional Networks}. In \bibinfo{booktitle}{\emph{{ICCV}}}.
\newblock


\bibitem[Tran et~al\mbox{.}(2018)]%
        {DBLP:conf/cvpr/TranWTRLP18}
\bibfield{author}{\bibinfo{person}{Du Tran}, \bibinfo{person}{Heng Wang},
  \bibinfo{person}{Lorenzo Torresani}, \bibinfo{person}{Jamie Ray},
  \bibinfo{person}{Yann LeCun}, {and} \bibinfo{person}{Manohar Paluri}.}
  \bibinfo{year}{2018}\natexlab{}.
\newblock \showarticletitle{A Closer Look at Spatiotemporal Convolutions for
  Action Recognition}. In \bibinfo{booktitle}{\emph{{CVPR}}}.
  \bibinfo{pages}{6450--6459}.
\newblock


\bibitem[Tu et~al\mbox{.}(2021a)]%
        {DBLP:journals/tip/TuWBAB21}
\bibfield{author}{\bibinfo{person}{Zhengzhong Tu}, \bibinfo{person}{Yilin
  Wang}, \bibinfo{person}{Neil Birkbeck}, \bibinfo{person}{Balu Adsumilli},
  {and} \bibinfo{person}{Alan~C. Bovik}.} \bibinfo{year}{2021}\natexlab{a}.
\newblock \showarticletitle{{UGC-VQA:} Benchmarking Blind Video Quality
  Assessment for User Generated Content}.
\newblock \bibinfo{journal}{\emph{{IEEE} TIP}}  \bibinfo{volume}{30}
  (\bibinfo{year}{2021}), \bibinfo{pages}{4449--4464}.
\newblock


\bibitem[Tu et~al\mbox{.}(2021b)]%
        {DBLP:journals/corr/abs-2101-10955}
\bibfield{author}{\bibinfo{person}{Zhengzhong Tu}, \bibinfo{person}{Xiangxu
  Yu}, \bibinfo{person}{Yilin Wang}, \bibinfo{person}{Neil Birkbeck},
  \bibinfo{person}{Balu Adsumilli}, {and} \bibinfo{person}{Alan~C. Bovik}.}
  \bibinfo{year}{2021}\natexlab{b}.
\newblock \showarticletitle{{RAPIQUE:} Rapid and Accurate Video Quality
  Prediction of User Generated Content}.
\newblock \bibinfo{journal}{\emph{CoRR}}  \bibinfo{volume}{abs/2101.10955}
  (\bibinfo{year}{2021}).
\newblock


\bibitem[Wang et~al\mbox{.}(2020)]%
        {DBLP:journals/corr/abs-2006-04768}
\bibfield{author}{\bibinfo{person}{Sinong Wang}, \bibinfo{person}{Belinda~Z.
  Li}, \bibinfo{person}{Madian Khabsa}, \bibinfo{person}{Han Fang}, {and}
  \bibinfo{person}{Hao Ma}.} \bibinfo{year}{2020}\natexlab{}.
\newblock \showarticletitle{Linformer: Self-Attention with Linear Complexity}.
\newblock \bibinfo{journal}{\emph{CoRR}}  \bibinfo{volume}{abs/2006.04768}
  (\bibinfo{year}{2020}).
\newblock


\bibitem[Wang et~al\mbox{.}(2021b)]%
        {DBLP:journals/corr/abs-2102-12122}
\bibfield{author}{\bibinfo{person}{Wenhai Wang}, \bibinfo{person}{Enze Xie},
  \bibinfo{person}{Xiang Li}, \bibinfo{person}{Deng{-}Ping Fan},
  \bibinfo{person}{Kaitao Song}, \bibinfo{person}{Ding Liang},
  \bibinfo{person}{Tong Lu}, \bibinfo{person}{Ping Luo}, {and}
  \bibinfo{person}{Ling Shao}.} \bibinfo{year}{2021}\natexlab{b}.
\newblock \showarticletitle{Pyramid Vision Transformer: {A} Versatile Backbone
  for Dense Prediction without Convolutions}.
\newblock \bibinfo{journal}{\emph{CoRR}}  \bibinfo{volume}{abs/2102.12122}
  (\bibinfo{year}{2021}).
\newblock


\bibitem[Wang et~al\mbox{.}(2018)]%
        {DBLP:conf/cvpr/0004GGH18}
\bibfield{author}{\bibinfo{person}{Xiaolong Wang}, \bibinfo{person}{Ross~B.
  Girshick}, \bibinfo{person}{Abhinav Gupta}, {and} \bibinfo{person}{Kaiming
  He}.} \bibinfo{year}{2018}\natexlab{}.
\newblock \showarticletitle{Non-Local Neural Networks}. In
  \bibinfo{booktitle}{\emph{{CVPR}}}. \bibinfo{pages}{7794--7803}.
\newblock


\bibitem[Wang and Adsumilli(2019)]%
        {DBLP:conf/mmsp/WangIA19}
\bibfield{author}{\bibinfo{person}{Yilin Wang} {and} \bibinfo{person}{Balu
  Adsumilli}.} \bibinfo{year}{2019}\natexlab{}.
\newblock \showarticletitle{YouTube {UGC} Dataset for Video Compression
  Research}. In \bibinfo{booktitle}{\emph{{MMSP}}}.
  \bibinfo{publisher}{{IEEE}}, \bibinfo{pages}{1--5}.
\newblock


\bibitem[Wang et~al\mbox{.}(2021a)]%
        {DBLP:conf/cvpr/WangKTYBAMY21}
\bibfield{author}{\bibinfo{person}{Yilin Wang}, \bibinfo{person}{Junjie Ke},
  \bibinfo{person}{Hossein Talebi}, \bibinfo{person}{Joong~Gon Yim},
  \bibinfo{person}{Neil Birkbeck}, \bibinfo{person}{Balu Adsumilli},
  \bibinfo{person}{Peyman Milanfar}, {and} \bibinfo{person}{Feng Yang}.}
  \bibinfo{year}{2021}\natexlab{a}.
\newblock \showarticletitle{Rich Features for Perceptual Quality Assessment of
  {UGC} Videos}. In \bibinfo{booktitle}{\emph{{CVPR}}}.
  \bibinfo{pages}{13435--13444}.
\newblock


\bibitem[Wang et~al\mbox{.}(2017)]%
        {DBLP:conf/cvpr/WangLWY17}
\bibfield{author}{\bibinfo{person}{Yunbo Wang}, \bibinfo{person}{Mingsheng
  Long}, \bibinfo{person}{Jianmin Wang}, {and} \bibinfo{person}{Philip~S. Yu}.}
  \bibinfo{year}{2017}\natexlab{}.
\newblock \showarticletitle{Spatiotemporal Pyramid Network for Video Action
  Recognition}. In \bibinfo{booktitle}{\emph{{CVPR}}}.
  \bibinfo{publisher}{{IEEE} Computer Society}, \bibinfo{pages}{2097--2106}.
\newblock


\bibitem[Wang et~al\mbox{.}(2000)]%
        {DBLP:conf/icip/WangBE00}
\bibfield{author}{\bibinfo{person}{Zhou Wang}, \bibinfo{person}{Alan~C. Bovik},
  {and} \bibinfo{person}{Brian~L. Evans}.} \bibinfo{year}{2000}\natexlab{}.
\newblock \showarticletitle{Blind Measurement of Blocking Artifacts in Images}.
  In \bibinfo{booktitle}{\emph{{ICIP}}}. \bibinfo{publisher}{{IEEE}},
  \bibinfo{pages}{981--984}.
\newblock


\bibitem[Wu et~al\mbox{.}(2022)]%
        {DBLP:journals/corr/abs-2206-09853}
\bibfield{author}{\bibinfo{person}{Haoning Wu}, \bibinfo{person}{Chaofeng
  Chen}, \bibinfo{person}{Liang Liao}, \bibinfo{person}{Jingwen Hou},
  \bibinfo{person}{Wenxiu Sun}, \bibinfo{person}{Qiong Yan}, {and}
  \bibinfo{person}{Weisi Lin}.} \bibinfo{year}{2022}\natexlab{}.
\newblock \showarticletitle{DisCoVQA: Temporal Distortion-Content Transformers
  for Video Quality Assessment}.
\newblock \bibinfo{journal}{\emph{CoRR}}  \bibinfo{volume}{abs/2206.09853}
  (\bibinfo{year}{2022}).
\newblock


\bibitem[Xie et~al\mbox{.}(2018)]%
        {DBLP:conf/eccv/XieSHTM18}
\bibfield{author}{\bibinfo{person}{Saining Xie}, \bibinfo{person}{Chen Sun},
  \bibinfo{person}{Jonathan Huang}, \bibinfo{person}{Zhuowen Tu}, {and}
  \bibinfo{person}{Kevin Murphy}.} \bibinfo{year}{2018}\natexlab{}.
\newblock \showarticletitle{Rethinking Spatiotemporal Feature Learning:
  Speed-Accuracy Trade-offs in Video Classification}. In
  \bibinfo{booktitle}{\emph{ECCV}}. \bibinfo{pages}{318--335}.
\newblock


\bibitem[Xing et~al\mbox{.}(2021)]%
        {DBLP:journals/corr/abs-2108-09635}
\bibfield{author}{\bibinfo{person}{Fengchuang Xing},
  \bibinfo{person}{Yuan{-}Gen Wang}, \bibinfo{person}{Hanpin Wang},
  \bibinfo{person}{Leida Li}, {and} \bibinfo{person}{Guopu Zhu}.}
  \bibinfo{year}{2021}\natexlab{}.
\newblock \showarticletitle{StarVQA: Space-Time Attention for Video Quality
  Assessment}.
\newblock \bibinfo{journal}{\emph{CoRR}}  \bibinfo{volume}{abs/2108.09635}
  (\bibinfo{year}{2021}).
\newblock


\bibitem[Xu et~al\mbox{.}(2021)]%
        {DBLP:conf/mm/XuLZZW021}
\bibfield{author}{\bibinfo{person}{Jiahua Xu}, \bibinfo{person}{Jing Li},
  \bibinfo{person}{Xingguang Zhou}, \bibinfo{person}{Wei Zhou},
  \bibinfo{person}{Baichao Wang}, {and} \bibinfo{person}{Zhibo Chen}.}
  \bibinfo{year}{2021}\natexlab{}.
\newblock \showarticletitle{Perceptual Quality Assessment of Internet Videos}.
  In \bibinfo{booktitle}{\emph{{ACM} Multimedia}}. \bibinfo{publisher}{{ACM}},
  \bibinfo{pages}{1248--1257}.
\newblock


\bibitem[Yang et~al\mbox{.}(2020)]%
        {DBLP:journals/corr/abs-2006-15489}
\bibfield{author}{\bibinfo{person}{Ceyuan Yang}, \bibinfo{person}{Yinghao Xu},
  \bibinfo{person}{Bo Dai}, {and} \bibinfo{person}{Bolei Zhou}.}
  \bibinfo{year}{2020}\natexlab{}.
\newblock \showarticletitle{Video Representation Learning with Visual Tempo
  Consistency}.
\newblock \bibinfo{journal}{\emph{CoRR}}  \bibinfo{volume}{abs/2006.15489}
  (\bibinfo{year}{2020}).
\newblock


\bibitem[Ye et~al\mbox{.}(2012)]%
        {DBLP:conf/cvpr/YeKKD12}
\bibfield{author}{\bibinfo{person}{Peng Ye}, \bibinfo{person}{Jayant Kumar},
  \bibinfo{person}{Le Kang}, {and} \bibinfo{person}{David~S. Doermann}.}
  \bibinfo{year}{2012}\natexlab{}.
\newblock \showarticletitle{Unsupervised feature learning framework for
  no-reference image quality assessment}. In
  \bibinfo{booktitle}{\emph{{CVPR}}}. \bibinfo{publisher}{{IEEE} Computer
  Society}, \bibinfo{pages}{1098--1105}.
\newblock


\bibitem[Ying et~al\mbox{.}(2021)]%
        {DBLP:conf/cvpr/YingMGB21}
\bibfield{author}{\bibinfo{person}{Zhenqiang Ying}, \bibinfo{person}{Maniratnam
  Mandal}, \bibinfo{person}{Deepti Ghadiyaram}, {and} \bibinfo{person}{Alan~C.
  Bovik}.} \bibinfo{year}{2021}\natexlab{}.
\newblock \showarticletitle{Patch-VQ: 'Patching Up' the Video Quality Problem}.
  In \bibinfo{booktitle}{\emph{{CVPR}}}. \bibinfo{publisher}{Computer Vision
  Foundation / {IEEE}}, \bibinfo{pages}{14019--14029}.
\newblock


\bibitem[You and Korhonen(2019)]%
        {DBLP:conf/icip/YouK19}
\bibfield{author}{\bibinfo{person}{Junyong You} {and} \bibinfo{person}{Jari
  Korhonen}.} \bibinfo{year}{2019}\natexlab{}.
\newblock \showarticletitle{Deep Neural Networks for No-Reference Video Quality
  Assessment}. In \bibinfo{booktitle}{\emph{{ICIP}}}.
  \bibinfo{publisher}{{IEEE}}, \bibinfo{pages}{2349--2353}.
\newblock


\bibitem[Yuan et~al\mbox{.}(2021)]%
        {Yuan_2021_ICCV}
\bibfield{author}{\bibinfo{person}{Kun Yuan}, \bibinfo{person}{Shaopeng Guo},
  \bibinfo{person}{Ziwei Liu}, \bibinfo{person}{Aojun Zhou},
  \bibinfo{person}{Fengwei Yu}, {and} \bibinfo{person}{Wei Wu}.}
  \bibinfo{year}{2021}\natexlab{}.
\newblock \showarticletitle{Incorporating Convolution Designs Into Visual
  Transformers}. In \bibinfo{booktitle}{\emph{{ICCV}}}.
  \bibinfo{publisher}{{IEEE}}, \bibinfo{pages}{579--588}.
\newblock


\bibitem[Zhang et~al\mbox{.}(2018)]%
        {DBLP:conf/cvpr/ZhangIESW18}
\bibfield{author}{\bibinfo{person}{Richard Zhang}, \bibinfo{person}{Phillip
  Isola}, \bibinfo{person}{Alexei~A. Efros}, \bibinfo{person}{Eli Shechtman},
  {and} \bibinfo{person}{Oliver Wang}.} \bibinfo{year}{2018}\natexlab{}.
\newblock \showarticletitle{The Unreasonable Effectiveness of Deep Features as
  a Perceptual Metric}. In \bibinfo{booktitle}{\emph{{CVPR}}}.
  \bibinfo{pages}{586--595}.
\newblock


\bibitem[Zhang et~al\mbox{.}(2019)]%
        {DBLP:journals/tcsv/ZhangGHLH19}
\bibfield{author}{\bibinfo{person}{Yu Zhang}, \bibinfo{person}{Xinbo Gao},
  \bibinfo{person}{Lihuo He}, \bibinfo{person}{Wen Lu}, {and}
  \bibinfo{person}{Ran He}.} \bibinfo{year}{2019}\natexlab{}.
\newblock \showarticletitle{Blind Video Quality Assessment With Weakly
  Supervised Learning and Resampling Strategy}.
\newblock \bibinfo{journal}{\emph{{IEEE} TCSVT}} \bibinfo{volume}{29},
  \bibinfo{number}{8} (\bibinfo{year}{2019}), \bibinfo{pages}{2244--2255}.
\newblock


\bibitem[Zhao et~al\mbox{.}(2023b)]%
        {Zhao_2023_CVPR}
\bibfield{author}{\bibinfo{person}{Kai Zhao}, \bibinfo{person}{Kun Yuan},
  \bibinfo{person}{Ming Sun}, \bibinfo{person}{Mading Li}, {and}
  \bibinfo{person}{Xing Wen}.} \bibinfo{year}{2023}\natexlab{b}.
\newblock \showarticletitle{Quality-Aware Pre-Trained Models for Blind Image
  Quality Assessment}. In \bibinfo{booktitle}{\emph{CVPR}}.
  \bibinfo{publisher}{{IEEE} Computer Society}, \bibinfo{pages}{22302--22313}.
\newblock


\bibitem[Zhao et~al\mbox{.}(2023a)]%
        {Zhao_2023_CVPR_Zoom}
\bibfield{author}{\bibinfo{person}{Kai Zhao}, \bibinfo{person}{Kun Yuan},
  \bibinfo{person}{Ming Sun}, {and} \bibinfo{person}{Xing Wen}.}
  \bibinfo{year}{2023}\natexlab{a}.
\newblock \showarticletitle{Zoom-VQA: Patches, Frames and Clips Integration for
  Video Quality Assessment}. In \bibinfo{booktitle}{\emph{CVPR Workshops}}.
  \bibinfo{publisher}{{IEEE} Computer Society}, \bibinfo{pages}{1302--1310}.
\newblock


\bibitem[Zhou and Zhang(2014)]%
        {zhou2014no}
\bibfield{author}{\bibinfo{person}{Luo-yu Zhou} {and}
  \bibinfo{person}{Zheng-bing Zhang}.} \bibinfo{year}{2014}\natexlab{}.
\newblock \showarticletitle{No-reference image quality assessment based on
  noise, blurring and blocking effect}.
\newblock \bibinfo{journal}{\emph{Optik}} \bibinfo{volume}{125},
  \bibinfo{number}{19} (\bibinfo{year}{2014}), \bibinfo{pages}{5677--5680}.
\newblock


\end{thebibliography}


\end{document}